%% file: acl2023.tex
\pdfoutput=1

\documentclass[11pt]{article}

\usepackage[]{ACL2023}

\usepackage{times}
\usepackage{latexsym}

\usepackage[T1]{fontenc}

\usepackage[utf8]{inputenc}

\usepackage{microtype}

\usepackage{inconsolata}

\usepackage{graphicx}
\usepackage{color}
\usepackage{makecell}
\usepackage{multirow}
\usepackage{amssymb}
\usepackage{booktabs}
\usepackage{tabularx}
\usepackage{tikz}
\usepackage[normalem]{ulem}
\usepackage{tcolorbox}
\usepackage{amsmath}
\usetikzlibrary{cd}
\usepackage{bm}
\usepackage{arydshln}
\usepackage{xcolor}
\usepackage{algpseudocode}
\usepackage{bbding}
\usepackage{dcolumn}
\usepackage{utfsym}
\usepackage{enumitem}
\usepackage[linesnumbered,ruled,vlined]{algorithm2e}

\title{Enhancing Language Representation with Constructional Information for Natural Language Understanding}

\author{Lvxiaowei Xu, Jianwang Wu, Jiawei Peng, Zhilin Gong, Ming Cai$^{*}$, Tianxiang Wang \\
        Department of Computer Science and Technology, Zhejiang University \\
        \texttt{\{xlxw, wujw, pengjw, zhilingong, cm, wang\_tx\}@zju.edu.cn}}

\newcommand\blfootnote[1]{%
  \begingroup
  \renewcommand\thefootnote{}\footnote{#1}%
  \addtocounter{footnote}{-1}%
  \endgroup
}

\begin{document}
\maketitle

\input{Chapters/abstract}

\input{Chapters/introduction}

\input{Chapters/construction}

\input{Chapters/method}

\input{Chapters/experiment}

\input{Chapters/relatedwork}

\input{Chapters/conclusion}

\input{Chapters/limitation}

\input{Chapters/acknowledgement}

\bibliography{anthology,custom}
\bibliographystyle{acl_natbib}

\input{Chapters/appendix}
\end{document}

%% file: Chapters/abstract.tex
\begin{abstract}
Natural language understanding (NLU) is an essential branch of natural language processing, which relies on representations generated by pre-trained language models (PLMs). However, PLMs primarily focus on acquiring lexico-semantic information, while they may be unable to adequately handle the meaning of constructions. To address this issue, we introduce construction grammar (CxG), which highlights the pairings of form and meaning, to enrich language representation. We adopt usage-based construction grammar as the basis of our work, which is highly compatible with statistical models such as PLMs. Then a HyCxG framework is proposed to enhance language representation through a three-stage solution. First, all constructions are extracted from sentences via a slot-constraints approach. As constructions can overlap with each other, bringing redundancy and imbalance, we  formulate the conditional max coverage problem for selecting the discriminative constructions. Finally, we propose a relational hypergraph attention network to acquire representation from constructional information by capturing high-order word interactions among constructions. Extensive experiments demonstrate the superiority of the proposed model on a variety of NLU tasks.
\end{abstract}

%% file: Chapters/introduction.tex
\section{Introduction}
\label{sec:introduction}

Recent progress in natural language processing relies on pre-trained language models (PLMs) \cite{devlin2019bert,liu2019roberta,raffel2020exploring,brown2020language,raffel2020exploring,he2020deberta}, which generate word representations by capturing syntactic and semantic features from their context. Typically, PLMs (e.g., BERT) employ the Masked Language Model (MLM) as a pre-training objective, randomly masking some tokens from the input before predicting the original tokens.
\blfootnote{$^*$ Corresponding author.}
However, PLMs primarily focus on acquiring lexico-semantic information \cite{tayyar2020cxgbert} while the meaning is not attached to the words instantiating the construction, but rather to the abstract pattern itself \cite{weissweiler2023construction}. Learning the representation of more complex and abstract linguistic units (called constructions) can be more challenging, such as collocations, argument structures and sentence patterns. For example, the ditransitive construction \cite{goodberg1995constructions}, which involves the form ``Subject--Verb--Object1--Object2'', denotes the meaning of transfer or giving of an entity (Object2) from an agent (Subject) to a recipient (Object1), such as ``\emph{John gave Mary a book}''. Linguistic experiments have proved that constructions can substantially contribute to sentence comprehension in multiple languages, such as English \cite{bencini2000contribution} and German \cite{gries2005foreign}.

\input{Tables/exp-introduction}

As the meaning of a construction is assigned to a language pattern rather than specific words \cite{hoffmann2013oxford}, the constructional information is rarely acquired by MLM and requires large bulk training data, which may lead to failure in natural language understanding tasks with constrained low resource data. As shown in Table~\ref{tab:example-introduction}, we illustrate three examples from aspect-based sentiment analysis (ABSA) tasks. The attention scores are derived from BERT-SPC \cite{devlin2019bert}, a simple baseline by fine-tuning aspect-specific representation of BERT. In the first example, the model focuses mainly on the opinion word ``\emph{friendly}'' while ignoring the modality construction of ``\emph{should be}'', resulting in the misclassification. In the second case, ``\emph{too hard to}'' expresses a negation for the ``\emph{fancy food}''. However, the model ignores the collocation of ``\emph{too...to}'' structure, and the ``\emph{food}'' is wrongly considered to be positive. The third example illustrates that lack of understanding of conditional sentence (\emph{if}  clause) causes the model to make an incorrect prediction.

Observations from both linguistics and NLU standpoints motivate us to exploit construction grammar \citep[CxG;][]{goodberg1995constructions,goldberg2006constructions} as the inductive bias to complement language representation with constructional information. Construction grammar refers to a family of linguistic approaches which regard constructions as the basic unit of language. Composed of pairings of form and meaning, constructions pervade all levels of language along a continuum (e.g., morphology, lexicon and syntax).

However, when bridging the gap between CxG and NLU, we face three critical questions:
\vspace{-0.5em}
\begin{enumerate}[itemsep=2pt,parsep=0pt,label=(Q\arabic*)]
    \item \label{ques:1} Which CxG approach is applicable for NLU?
    
    \item \label{ques:2} How can typical constructions be identified?
    
    \item \label{ques:3} How can constructions be encoded?
\end{enumerate}
\vspace{-0.5em}

To answer \ref{ques:1}, we investigate different variants of CxG. Instead of formal methods (e.g., fluid construction grammar and embodied construction grammar), we adopt usage-based approaches which assume that grammar acquisition involves statistical inference. Therefore constructions can be induced as frequent instances of linguistic units. This assumption makes usage-based approaches highly compatible with statistical models such as PLMs \cite{kapatsinski2014grammar}. Specifically, we follow the efforts of \citet{dunn2019frequency} with a computationally slot-constraints approach that formulates constructions as the combinations of immutable components and fillable slots. As shown in Table~\ref{tab:example-introduction}, the modality construction ``NOUN--AUX--\emph{be}'' in the first example contains an immutable component ``\emph{be}'' and two fillable slots (NOUN and AUX) with a meaning of expressing advice or suggestion. 

As for \ref{ques:2}, from the usage-based perspective, constructions are often stored redundantly at different levels of language. Therefore constructions can overlap with each other, which results in redundancy and imbalance. Consequently, we formulate the problem of selecting discriminative constructions as multi-objective optimization for conditional maximum coverage. To alleviate the computational complexity, we adopt a heuristic search algorithm, i.e., simulated annealing (SA) to determine the composition of constructions.

In order to address \ref{ques:3}, we propose a relational hypergraph attention network (R-HGAT) to capture high-order word interactions inside the constructions, so as to acquire representation from constructional information. R-HGAT generalizes hypergraph attention network using its flexible labeled hyperedges. We refer to the entire framework which involves construction extraction, selection and encoding as hypergraph network of construction grammar (HyCxG)\footnote{Our code is publicly avaliable at \url{https://github.com/xlxwalex/HyCxG}.}.

Extensive experiments have been conducted to illustrate the superiority of HyCxG on NLU tasks, while multilingual experiments further indicate the constructional information is beneficial across languages. Additionally, based on the constructional representations acquired by our model, we conduct an empirical study of building a network of constructions, which provides meaningful implication and in turn contributes to usage-based CxG.

%% file: Tables/exp-introduction.tex
\newcommand{\myboxl}[1]{\tikz[baseline=(MeNode.base)]{\node[rectangle, minimum height=12pt,inner sep=1pt, fill=orange!15](MeNode){#1};}}
\newcommand{\myboxm}[1]{\tikz[baseline=(MeNode.base)]{\node[rectangle, minimum height=12pt,inner sep=1pt, fill=orange!25](MeNode){#1};}}
\newcommand{\myboxhl}[1]{\tikz[baseline=(MeNode.base)]{\node[rectangle, minimum height=12pt, inner sep=1pt, fill=orange!35](MeNode){#1};}}
\newcommand{\myboxh}[1]{\tikz[baseline=(MeNode.base)]{\node[rectangle, minimum height=12pt,inner sep=1pt, fill=orange!42](MeNode){#1};}}
\newcommand{\myboxhh}[1]{\tikz[baseline=(MeNode.base)]{\node[rectangle, minimum height=12pt,inner sep=1pt, fill=orange!60](MeNode){#1};}}

\begin{table}[t]
    \fontsize{9.5}{11}\selectfont
	\centering
	\begin{tabular}{p{4.5cm} p{2.5cm} }
		\toprule
	    \textbf{Sentence}&\textbf{Construction} \cr
		\midrule
		\myboxhl{The} \myboxhh{[staff]$_\text{N}$} should be \myboxl{a} bit \myboxl{more} \myboxh{friendly}.                 &NOUN--AUX--be \quad \quad (staff--should--be)\cr \hline
		The restaurants \myboxhl{try} too \myboxl{hard} to make \myboxh{fancy} \myboxhh{[food]$_\text{N}$}.                 &\rule[0pt]{0pt}{13pt}ADV--hard--to \quad \quad (too--hard--to)\cr \hline
	    I can \myboxl{understand} the \myboxm{prices} if it \myboxl{served} \myboxh{better} \myboxhh{[food]$_\text{N}$}.    &\rule[0pt]{0pt}{15pt}if--PRON--VERB \quad \quad (if--it--served)\cr
	\bottomrule
	\end{tabular}
	\caption{Visualization of attention scores from misclassified examples in aspect-based sentiment analysis. [$\cdot$]$_\text{N}$ refers to the  aspects with negative sentiment polarity.}
	\label{tab:example-introduction}
\end{table}

%% file: Chapters/construction.tex
\section{Construction Extraction and Selection}
\label{sec:cxg-selection}

In this section, we elaborate on the details of extraction \ref{ques:1} and selection of constructions \ref{ques:2}.

\subsection{Computational Construction Grammar}
\label{sec:construction-grammar}

Construction grammar is a branch of cognitive linguistics which assumes that syntax and patterns have specific meanings. A construction is a form-meaning pair whose structure varies in different level of abstractness, including partially or fully filled components and general linguistic patterns. According to CxG, speakers can recognize patterns after coming across them a certain number of times, comparable to merging n-grams (i.e., constructions) at different schematic or abstract levels \cite{goldberg2003constructions,goldberg2006constructions,tayyar2020cxgbert}. However, the methodology for acquiring the constructions is always labor-intensive and requires careful definition, while the computational generation of CxG is a relatively new research field.

As our answer to \ref{ques:1}, the usage-based approach is employed to obtain constructions following the work of \citet{dunn2019frequency}. They propose a grammar induction algorithm, which can learn the patterns of constructions with a slot-constraints approach from the corpus. Therefore, we adapt \citeposs{dunn2019frequency} system in our work for construction extraction.

Constructions are represented as a combination of slots and are separated by dashes (Table~\ref{tab:example-introduction}). The slot fillers are drawn from lexical units (words), semantic categories (the discrete semantic domains are clustered with fastText embeddings \cite{grave2018learning} via K-Means) and syntax \citep[Universal Part-of-Speech tags;][]{petrov2012universal}, which are progressively higher in abstraction level.
 
\subsection{Conditional Max Coverage Problem}
\label{sec:max-coverage}

Since constructions are often stored redundantly at different levels of abstractness, overlapping constructions can be captured by the grammar induction algorithm \cite{dunn2017computational,dunn2019frequency}. We summarize the phenomenon of overlap into two scenarios: \emph{Inclusion} and \emph{Intersection}. \emph{Inclusion} refers to the case where one construction is a subpart or instantiation of another construction, while \emph{Intersection} indicates that the constructions have some common slots. Redundancy and imbalance encoding problem can be brought by overlap phenomenon. The redundancy of the constructions can cause high computational complexity. And imbalanced construction distribution may introduce interference. This is due to the fact that the words in high dense parts are updated frequently, while this seldom happens in low dense parts. There is a virtual example to illustrate these phenomena in Figure~\ref{fig:method-hycxg}.

Therefore, the key issue is to select the discriminative constructions from all candidates \ref{ques:2}.  We formulate this as multi-objective optimization for conditional maximum coverage (Cond-MC) problem. Specifically, we seek an optimal subset $\mathcal{C}_O=\{c_1, c_2, \cdots, c_m\}$ from the set $\mathcal{C}=\{c_1, c_2, \cdots, c_n\}$ containing all $n$ constructions. The constructions in $\mathcal{C}_O$ are supposed to reach the following three objectives:
\vspace{-0.5em}
\begin{enumerate} [itemsep=1pt,parsep=0pt]
    \item \label{mcobj:1} The constructions cover as many words as possible in sentences.
    \item \label{mcobj:2} Intersection among constructions is minimal.
    \item \label{mcobj:3} Constructions preferably contain more slots from concrete levels.
\end{enumerate}
\vspace{-0.5em}

Objectives \ref{mcobj:1} and \ref{mcobj:2} spread the constructions throughout the entire sentence to ensure balanced distribution, while the more discriminative constructions can be selected by objective \ref{mcobj:3}. To unify the objective functions for optimization, we formalize each objective as a specific score function and weight them together to generate the total score. Therefore, Cond-MC can be converted to the problem of maximizing the score for $\mathcal{C}_O$. We define each construction $c_i$ as a set of $r$ slots $c_i=\{w_1, w_2, \cdots, w_r\}$, while $\operatorname{D}(\cdot)$ is utilized to compute the number of elements in a set. Then the score function of objectives \ref{mcobj:1} and \ref{mcobj:2} can be formulated as $\operatorname{D}(c_1\cup\cdots\cup c_m)$ and $\operatorname{D}(c_1\cap\cdots\cap c_m)$, with $S_{\text{ob1}}$ and $S_{\text{ob2}}$ referring to them respectively. And the score function of objective \ref{mcobj:3} is written as:
\begin{equation}
    \label{eq:object-3}
    S_{\text{ob3}} = \sum_{i=1}^m \sum_{j=1}^r \frac{s_r}{r}, s_r \in \{s_{\text{syn}}, s_{\text{sem}}, s_{\text{lex}}\}
\end{equation}
where three scores $s_{\text{syn}}$, $s_{\text{sem}}$ and $s_{\text{lex}}$ are assigned for slots in syntax, semantic and lexical level. The score of each construction is calculated by averaging the slot scores, while $S_{\text{ob3}}$ computes the total score of all constructions in $\mathcal{C}_O$. Then Cond-MC can be formulated as:
\begin{equation}
    \label{eq:cond-mc}
    \left\{\begin{aligned}
        \max \quad &w_1S_{\text{ob1}} - w_2S_{\text{ob2}} + w_3S_{\text{ob3}} \\
        \mbox{s.t.} \quad &\mathcal{C}_O \subset \mathcal{C}
    \end{aligned} \right.
\end{equation}
where $w_1$, $w_2$ and $w_3$ are the weighting factors for balancing the scores of three objectives.

\input{Algorithms/algo-condmc}
\subsection{Solution for Cond-MC Problem}
\label{sec:simu-anneal}

As we formulate a multi-objective optimization problem for selecting typical constructions, it is complicated to solve Cond-MC since it is an NP problem. When the solution space is large, i.e., there is a significant amount of constructions in the sentence, which leads to an unacceptable computational complexity. To alleviate this issue, we employ the Simulated Annealing \citep[SA;][]{kirkpatrick1983optimization} algorithm in heuristic search of $\mathcal{C}_O$, where the procedure is stated in Algorithm~\ref{algo:cond-mc}.

We define $\mathcal{C}_O$ as a binary set to indicate whether a construction is selected. \textsc{RandFlip}($\cdot$) is applied to reverse $t$ elements stochastically, while step $k$ is inversely related to temperature $T$. \textsc{SCORE($\cdot$)} is the overall score function in Equation~\ref{eq:cond-mc}. Besides, we utilize \textsc{Rand()} to compare with the transition probability $P$ to estimate whether to accept the new solution under metropolis criterion (i.e., accepting a new solution with a certain probability to increase the perturbation). A cooling schedule is used to control the evolution of the new solution with a reduced temperature function \textsc{Cool}($\cdot$) to regulate the temperature in an exponential form.

%% file: Algorithms/algo-condmc.tex
\begin{algorithm}[t]
    \fontsize{10.1}{12}\selectfont
    \SetAlgoLined
    \DontPrintSemicolon
	\caption{SA algorithm for Cond-MC}
	\label{algo:cond-mc}
	\KwIn{The set of all constructions $\mathcal{C}$.}
	\KwOut{The optimal set $\mathcal{C}_O$.}
	\SetKwFunction{Flip}{RANDFLIP}
	\SetKwFunction{Score}{SCORE}
	\SetKwFunction{Rand}{RAND}
	\SetKwFunction{Cool}{COOL}
    \SetKwProg{Fn}{Function}{:}{}
	
	Initialize the feasible solutions $\mathcal{C}_O \gets \mathcal{C}_f$\;
	Initialize temperature $T \gets T_{0}$, step $k \gets 0$\;
	\While{$k$ \textless $k_{\text{max}}$}{
	    $T \gets$ \Cool{$T$, $k$, $k_{\text{max}}$}\;
	    $\mathcal{C}_k \gets$ \Flip{$\mathcal{C}_O$, $T$}\;
	    $dE \gets$ \Score{$\mathcal{C}_k$} $-$ \Score{$\mathcal{C}_O$}\;
	    \If{$dE > 0$ \textbf{or} \Rand{} $\leq P(dE, T)$ }{
	        $\mathcal{C}_O \gets \mathcal{C}_k$\;
	    }
	    $k \gets k + 1$\;
	}
	\textbf{return} $\mathcal{C}_O$ \;
	
\end{algorithm}

%% file: Chapters/method.tex
\input{Figures/method-hycxg}
\section{Model}
\label{sec:model}
After extracting and determining the optimal set for constructions, the next challenge is to encode constructions in an effective way \ref{ques:3}. In this section, we first introduce hypergraph for modelling complex data correlation within constructions. Then a relational hypergraph attention network (R-HGAT) is proposed to encode the high-order word interactions inside each construction. Finally, enhanced by the constructional information, the language representation is ready for NLU tasks.

\subsection{Construction Hypergraph Generation}
Since the associations within constructions are not only dyadic, but could be triadic, tetradic or even higher-order, we adopt hypergraph for modelling such data correlation. Hypergraph is a generalization of conventional graph that the hyperedge can connect arbitrary number of nodes. A hypergraph can be defined as $\mathcal{G}=\{ \mathcal{V}, \mathcal{E}\}$, which contains a node set $\mathcal{V}=\{v_1, v_2, \cdots, v_m\}$ and a hyperedge set $\mathcal{E}=\{e_1, e_2, \cdots, e_n\}$. As each hyperedge can connect two or more nodes, the hypergraph $\mathcal{G}$ can be denoted by an incidence matrix $\mathbf{H} \in \mathbb{R}^{|\mathcal{V}|\times |\mathcal{E}|}$, with entries being defined as:
\begin{equation}
\label{eq:incidence-mat}
\mathbf{H}_{i j}=\left\{\begin{array}{l}
1, \quad \text { if } v_i \in e_j \\
0, \quad \text { if } v_i \notin e_j
\end{array}\right.
\end{equation}
Given a sentence with $m$ words (i.e., nodes in $\mathcal{G}$), we can model high-order word interactions through $n$ typical constructions. In specific,  each construction is regarded as a hyperedge, while the words contained in the construction are considered as the member nodes of certain hyperedges. Thus, we can generate the hypergraph for each sentence as illustrated in Figure~\ref{fig:method-hycxg}.

\subsection{Hypergraph Attention Network}

Different from the simple graph that nodes are pairwise linked, the information of nodes cannot be directly aggregated from neighboring nodes in hypergraph. Thus, hypergraph attention network (HGATT) is employed to learn the node representations, which separates the computation procedure into two steps, i.e., node aggregation and hyperedge aggregation. It first aggregates the information of nodes for generating the hyperedge representation. Then the information is fused back to the nodes from hyperedges via hyperedge aggregation. In general, the procedure is formulated as follows:
\begin{equation}
\begin{aligned}
\mathbf{g}_j^l & \gets \operatorname{AGGR}_{\text {Node}}\left(\left\{\mathbf{h}_s^{l-1} \mid \forall v_s \in e_j\right\}\right) \\
\mathbf{h}_i^l & \gets \operatorname{AGGR}_{Edge}\left(\mathbf{h}_i^{l-1},\left\{\mathbf{g}_j^l \mid \forall e_j \in \mathcal{E}_i\right\}\right)
\end{aligned}
\end{equation}
where $\operatorname{AGGR}_{Node}$ and $ \operatorname{AGGR}_{\text {Edge}}$ denote the two-step aggregation functions for nodes and hyperedges. $\mathbf{h}_i^l$ and $\mathbf{g}_j^l$ denote the representations of node $v_i$ and hyperedge $e_j$ in $l$-th layer, while $\mathcal{E}_i$ is the set of hyperedges connected to node $v_i$.

The HGATT is mainly implemented based on graph attention mechanism \cite{velivckovic2017graph}, such as HyperGAT \cite{ding2020less}. However, this attention mechanism applies the same weight matrices for different types of information within hyperedges, which inhibits the ability for the models to capture high-order relations in hypergraphs \cite{fan2021heterogeneous}.

\subsection{R-HGAT Network}

To tackle the limitation of HGATT and exploit the information of constructions, we propose a relational hypergraph attention network (R-HGAT). The mutual attention mechanism, i.e., node-level attention and edge-level attention, is adopted to aggregate the information from nodes and hyperedges. The entire architecture is described as follows:

\paragraph{Node-level attention.} Node-level attention is utilized to encode representation of hyperedges that aggregates information from the nodes. Given the node $v_i$ and the set of hyperedges $\mathcal{E}_i$ that connect to $v_i$, we first embed each hyperedge (i.e., construction) into vector space $\mathbf{z}_j$ by looking up an embedding matrix $E_c \in \mathbb{R}^{|\mathcal{V}|\times d}$, while $|\mathcal{V}|$ is the size of construction set within a certain language and $d$ refers to the dimension of the vectors. After that, the attention mechanism is applied to compute the importance score for spotlighting the nodes that are crucial to the hyperedge according to $\mathbf{z}_j$. Then the aggregation procedure can be formulated as: 
\begin{equation}
\label{eq:node-attn}
\mathbf{g}_j^l=\sum_{v_s \in e_j} \alpha_{js} W_n \mathbf{h}_s^{l-1}
\end{equation}
where $W_n$ is the weight matrix and $\alpha_{js}$ is the attention score between the representation of node $\mathbf{h}_s^{l-1}$ and hyperedge $\mathbf{z}_j$  that can be computed by: 
\begin{equation}
\begin{aligned}
\label{eq:node-attn-score}
\mathbf{r}_{js}^{l-1} &= \sigma((W_c \mathbf{z}_j)^{\mathrm{T}}W_s \mathbf{h}_s^{l-1})\\
\alpha_{j s} & =\frac{\exp (\mathbf{r}_{js}^{l-1})}{\sum_{v_s \in e_j} \exp (\mathbf{r}_{js}^{l-1})}
\end{aligned}
\end{equation}
where $W_c$ and $W_s$ are trainable matrices, while $\sigma(\cdot)$ is the non-linear activation function (e.g., ReLU).

In particular, the information of constructions is injected to hyperedges with trainable matrix $W_g$ as:
\begin{equation}
    \label{eq:inject-cxg-inform}
    \mathbf{g}_j^{'l}=\mathbf{g}_j^{l} + W_g \mathbf{z}_j
\end{equation}
\paragraph{Edge-level attention.}  As an inverse procedure, we fuse the information of the hyperedges back to each node via the edge-level attention. Formally:
\begin{equation}
\begin{aligned}
\label{eq:edge-attn}
\mathbf{h}_i^l&=\sum_{e_j \in \mathcal{E}_i} \beta_{ij} W_e \mathbf{g}_j^{'l}\\
\mathbf{t}_{ij}^{l-1} &= \sigma((W_o \mathbf{h}_i^{l-1})^{\mathrm{T}}W_r \mathbf{g}_j^{'l})\\
\beta_{ij} & =\frac{\exp (\mathbf{t}_{ij}^{l-1})}{\sum_{e_j \in \mathcal{E}_i} \exp (\mathbf{t}_{ij}^{l-1})}
\end{aligned}
\end{equation}
where $W_e$, $W_o$ and $W_r$ are trainable matrices. $\beta_{ij}$ is the attention score of hyperedge $e_j$ on node $v_i$.

After the enhanced representation $\mathbf{H}^l$ is acquired with mutual attention mechanism in R-HGAT network, we apply the feed-forward network (FFN) consisting of two fully-connected layers coupled with residual connections to generate final node representation $\mathbf{H}^{'l}$, it can be formulated as:
\begin{equation}
\label{eq:ffn}
\mathbf{H}^{'l} = \operatorname{LN}(\mathbf{H}^{l-1} + \sigma(W_2(W_1\mathbf{H}^{l}+b_1)+b_2))
\end{equation}
where $W_1$ and $W_2$ refer to trainable weight matrices, while $b_1$ and $b_2$ are the biases. Besides, $\operatorname{LN}(\cdot)$ denotes the layer normalization operation.

\subsection{Model Training}
To apply representations to downstream tasks, we utilize $\operatorname{average-pooling}$ to obtain the task-specific representation $\mathbf{h}_z$, which retains most of the information on node representations. For classification tasks (e.g., sentiment analysis and paraphrase tasks), $\mathbf{h}_z$ is passed to a fully connected $\operatorname{softmax}$ layer, where the objective function to be minimized is the cross-entropy loss. As for regression task (i.e., similarity task), mean squared error loss is adopted as the objective function for optimization.

%% file: Figures/method-hycxg.tex
\begin{figure*}[t]
	\centering
	\includegraphics[width=\textwidth]{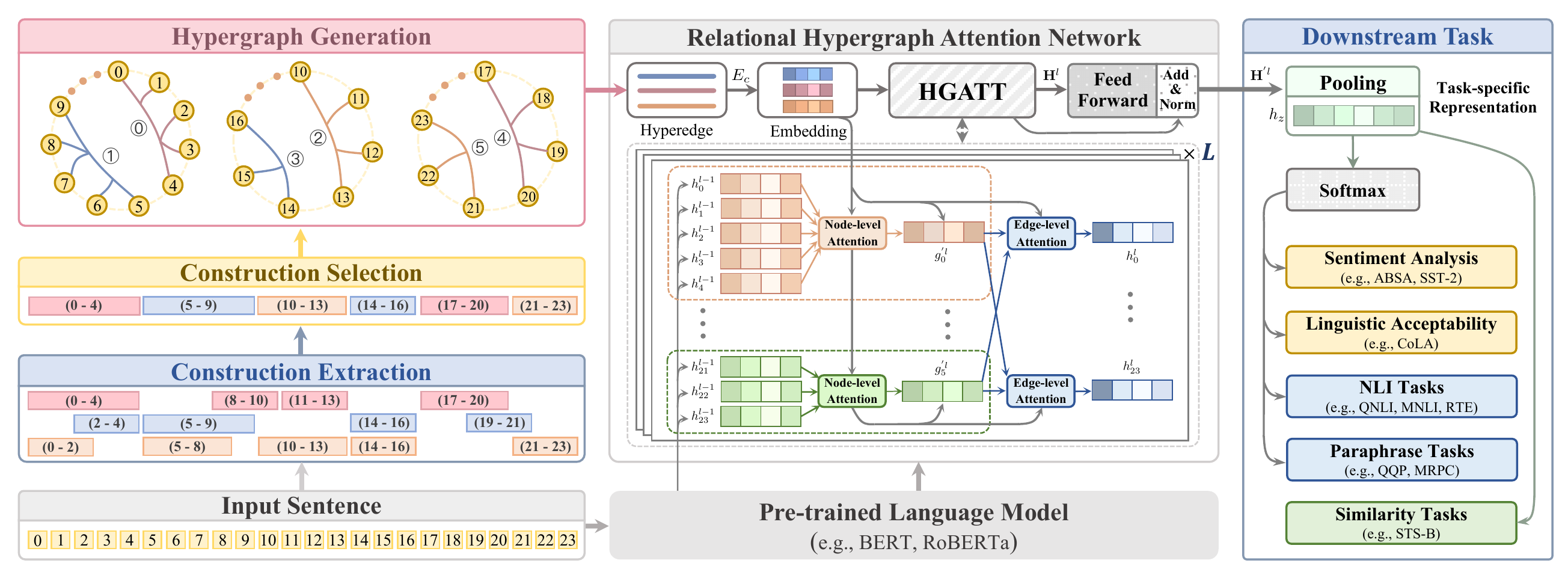} 
	\caption{Overview of the proposed HyCxG framework. For illustration, we show the entire system with a virtual example. In construction extraction module, CxG (5-8) is a subpart of CxG (5-9), which demonstrates the \emph{inclusion} relation of the overlap phenomenon, while CxG (5-9) and CxG (8-10) are \emph{intersection} relations.}
	\label{fig:method-hycxg}
\end{figure*}

%% file: Chapters/experiment.tex
\section{Experiments}
\label{sec:experiments}

\input{Tables/exp-absa}
\subsection{Experiments Setup}
\label{sec:exp-setup}

\paragraph{Datasets.} Experiments are conducted on a variety of NLU tasks for evaluation. We adopt ABSA datasets from SemEval and MAMS \cite{pontiki2014semeval,pontiki2015semeval,pontiki2016semeval,jiang2019challenge} as well as GLUE \cite{wang2018glue} benchmark to demonstrate the effectiveness of our HyCxG. GLUE contains a broad range of datasets, including natural language inference (MNLI, QNLI, RTE), sentence-level sentiment analysis (SST-2), paraphrase and similarity (MRPC, QQP, STS-B), and linguistic acceptability (CoLA). The detailed statistics of these datasets are provided in Appendix~\ref{appendix:data-statistic}.

\paragraph{Implementation.} To solve Cond-MC problem, we set the weight scores to 1.0, 1.2 and 1.5 for $s_{\text{syn}}$, $s_{\text{sem}}$ and $s_{\text{lex}}$ (Equation~\ref{eq:object-3}), while $w_1$, $w_2$ and $w_3$ are set to 1.0, 0.4 and 0.3 (Equation~\ref{eq:cond-mc}). Meanwhile, the \textsc{COOL}($\cdot$) function is formulated as $T_{0}\cdot a^k$, while $a$ is the factor calculated as $-\ln(T_{0}/T_{f})$.

HyCxG is optimized with AdamW \cite{loshchilov2018decoupled} optimizer. The optimal hyper-parameters are selected when the model achieves the highest performances on the development set via grid search. More detailed setups are described in Appendix~\ref{sec:appendix-hyper-params}, and the pre-trained model weights are obtained from Hugging Face \cite{wolf2020transformers}.

\paragraph{Baselines.}  We compare HyCxG with BERT-based state-of-the-art baselines on ABSA datasets: (1) \textbf{BERT-SPC} denotes the fine-tuning BERT with aspect-specific representation. (2) \textbf{LCFS-ASC} \cite{phan2020modelling} employs the syntactic distance to alleviate the interference from unrelated words. (3) \textbf{R-GAT} \cite{wang2020relational} utilizes a relational graph attention network to encode the pruned dependency tree. (4) \textbf{KumaGCN} \cite{chen2020inducing} synthesizes the dependency tree and latent graphs to enhance representation. (5) \textbf{DGEDT} \cite{tang2020dependency} provides a dependency enhanced dual-transformer for classification. (6) \textbf{DualGCN} \cite{li2021dualgraph} links the syntactic structure and semantic relevance to generate features. (7) \textbf{dotGCN} \cite{chen2022discrete} builds the induced tree with syntactic distances to encode opinion information. 

For other NLU tasks from the GLUE benchmark, we compare HyCxG with the base and large versions of the RoBERTa \cite{liu2019roberta} model to demonstrate the performance improvement.

\subsection{Experimental Results}
\label{sec:exp-results}

\input{Tables/exp-glue}

\paragraph{Results on ABSA tasks.} Experiments on ABSA tasks are conducted firstly, which are intuitive to analyze the function of constructions with aspects. Meanwhile, the results can be extended to multilingual settings as well as accessing to case studies. The model performances are shown in Table~\ref{tab:absa-exp-result}, from which several observations can be obtained. 

First, compared to BERT-SPC model, most of syntax-based baselines achieve higher performances on five datasets, since they can alleviate interference introduced and build the relationship between aspects and their corresponding opinion words via synthesizing syntactic knowledge (e.g., dependency tree). Second, the baseline models (i.e., DualGCN and dotGCN) with more information, such as semantic and opinion representations, gain better performances than others.

Third, our HyCxG significantly outperforms all baselines with constructional information incorporated, which demonstrates the effectiveness of HyCxG. Furthermore, the perspective that construction grammar can contribute to sentence comprehension for NLU tasks can also be substantiated. 

\paragraph{Results on GLUE tasks.} As shown in Table~\ref{tab:experiment_glue}, we conduct performance comparison among HyCxG with the base (B) and large (L) version of RoBERTa on GLUE development sets. And the results on test set are reported via online leaderboard. 

The average results over tasks illustrate that our HyCxG$_{\text{B}}$ outperforms RoBERTa$_{\text{B}}$, and HyCxG$_{\text{L}}$ is also better than RoBERTa$_{\text{L}}$ on both development and test sets. It indicates the validity of HyCxG and the benefits of constructional information.

As a branch of grammar, HyCxG has huge improvement on CoLA task, which shows its potential capability for evaluating linguistic acceptability. Meanwhile, HyCxG also significantly outperforms RoBERTa on RTE and MRPC tasks. Especially on the MRPC task, HyCxG$_{\text{B}}$ even surpasses RoBERTa$_{\text{L}}$, which further validates the language representation can be enhanced by constructional information. Besides, HyCxG also boosts the performances on all other NLU tasks as well.

\input{Tables/exp-oml}

\subsection{Comparative Analysis}
\label{sec:exp-analysis}

\paragraph{Ablation study on model complexity.} Though the efficiency of our HyCxG model is illustrated from the results on NLU tasks, we conduct ablation study to further investigate whether the performance enhancement is due to an increase in model complexity. We implement OML and OLL+HyCxG to address this concern. As increasing the depth of the PLMs can improve performance, OML adds an additional transformer layer on BERT, while OLL+HyCxG removes the last layer of BERT in HyCxG.  Besides, we continue pre-training OML model (OML+Pretrain) on the massive corpus to achieve its optimal performance.

As shown in Table~\ref{tab:experiment-oml}, OML achieves higher results than BERT-SPC, indicating that more complicated models tend to yield better performance. OML+Pretrain performs better compared to OML, which indicates the effect of pre-training. However, OLL+HyCxG surpasses OML with lack of two transformer layers, while it also overtakes OML+Pretrain on Rest 14 and 16 datasets. Meanwhile, HyCxG significantly outperforms OML and OML+Pretrain. As the details of computation complexity comparison between HyCxG and other baselines shown in Appendix~\ref{sec:computation-complexity-app}, the results illustrate that HyCxG achieves the highest performance at a relatively low computational complexity. These evidences all prove the efficacy of HyCxG and constructional information. The ablation study on GLUE benchmark is conducted in Appendix~\ref{sec:ablation-on-glue-app}.

\input{Tables/exp-coverage}

\paragraph{Comparison of construction selection strategies.}

In section~\ref{sec:cxg-selection}, we formulate the Cond-MC problem for selecting typical constructions. Table~\ref{tab:experiment-coverage} illustrates the performances for different strategies.

The strategy of w/o Cond denotes that constructions cannot overlap with each other, while w/o Selection employs all constructions. As redundancy and imbalance can occur in w/o Selection and w/o Cond, w/o Cond performs better on Rest 15 and 16 datasets, while fails on Rest 14 and Lap 14 datasets compared to w/o Selection. In contrast, Cond-MC strategy consistently outperforms w/o Selection and w/o Cond on all datasets. The results illustrate the necessity of selecting typical constructions and the validity of Cond-MC strategy.

\input{Tables/exp-multilingual}

\paragraph{Multilingual results.} 
\label{sec:multilingual}
Since construction grammar is applicable to multilingual analysis and can be learned via a unified framework for different languages \cite{dunn2022exposure}, we can further discuss the performance improvements of our HyCxG model in multilingual settings. The ABSA datasets in SemEval 2016 \cite{pontiki2016semeval} with multiple languages are conducted for our experiments (statistics are shown in Table~\ref{tab:statistic-absa}). Meanwhile, we also adapt baselines to multilingual environment with their official implementations. For models that involve dependency tree parsing (e.g., R-GAT, DualGCN), we maintain the same parsing tools that they employ (spaCy or Stanford Parser \cite{manning2014stanford}). The results are shown in Table~\ref{tab:experiment-ml}. We can observe that the inclusion of syntactic information can also improve model performances on multilingual settings compared to BERT-SPC. DualGCN has relatively high performance on different languages, since it incorporates the semantic features via SemGCN to the syntactic foundation with regularizers. Moreover, HyCxG outperforms these baselines across all four languages, indicating that constructional information can also enhance semantic understanding for other languages.

\paragraph{More empirical studies.} We conduct experiments to investigate the potential capacity of HyCxG. First, experiments in Appendix~\ref{sec:counterfactual-pattern-app} demonstrate the superior ability of HyCxG in perceiving statements and patterns. Second, we conduct ablation study in Appendix~\ref{sec:appendix-component} to examine the benefits for each component in HyCxG. We also investigate the results on colloquial datasets in Appendix~\ref{sec:appendix-informal}. As the constructions are derived from a formal language corpus, which has a different register \cite{dunn2023exploring} than the colloquial dataset, some of the edge constructions are not extracted to enhance the representation, resulting in suboptimal performances. It illustrates the necessity of obtaining constructions on a diverse corpus. Besides, we present an approach in Appendix~\ref{sec:constructicon-app} to build the construction network with the representations of constructions acquired in HyCxG. It depicts the inheritance relations between constructions, which provides meaningful implications to usage-based CxG.

\subsection{Case Study}
\label{sec:case-study}

The case study is utilized to illustrate the mechanism inherent in HyCxG for NLU tasks. We first visualize the learned representations via t-SNE \cite{van2008visualizing} in Figure~\ref{fig:analysis-tsne}, while the representations of these five randomly selected constructions are obtained via applying average-pooling in BERT-SPC and HyCxG. The results demonstrate that the representations in HyCxG form clusters with distinct boundaries, while the representations in BERT-SPC are diffuse among different clusters. It proves that HyCxG can model higher-order interactions and synthesize constructional information.

Meanwhile, we visualize the attention scores to illustrate the benefits of constructional information in Figure~\ref{fig:analysis-attnvis}. As discussed in Section~\ref{sec:introduction}, BERT-SPC mainly focuses on the opinion word ``\emph{friendly}'', leading to misclassification. In contrast, HyCxG can produce correct prediction, since it captures the modality pattern (i.e., NOUN--AUX--\emph{be}).

\input{Figures/analysis-tsne}
\input{Figures/analysis-attnvis}

%% file: Tables/exp-absa.tex
\begin{table*}[!t]	
	\centering
 	\fontsize{9}{11}\selectfont
	\resizebox{1.0\linewidth}{!}{
	    \begin{tabular}{lcccccccccc}
		\toprule
		\multirow{2}{*}{\rule[0pt]{0pt}{12pt}\textbf{Model}}
	     & \multicolumn{2}{c}{\textbf{Rest14}} 
	     & \multicolumn{2}{c}{\textbf{Lap14}} 
	     & \multicolumn{2}{c}{\textbf{Rest15}}
	     & \multicolumn{2}{c}{\textbf{Rest16}}
	     & \multicolumn{2}{c}{\textbf{MAMS}} \cr \cline{2-11}\rule[0pt]{0pt}{11pt}
		 &\textbf{Acc}&\textbf{F$_1$}&\textbf{Acc}&\textbf{F$_1$}&\textbf{Acc}&\textbf{F$_1$}&\textbf{Acc}&\textbf{F$_1$}&\textbf{Acc}&\textbf{F$_1$}\cr
		\midrule
		BERT-SPC   &85.09&78.43&79.47&75.90&83.21&61.43&90.74&74.54&82.34&81.94\cr
		LCFS-ASC   &86.71&80.31&80.52&77.13&82.47&66.39&89.77&75.23&82.78&82.25\cr
		R-GAT      &86.60&81.35&78.21&74.07&84.32&68.47&91.56&75.85&84.13&83.78\cr 
		KumaGCN    &86.43&80.30&81.98&78.81&\textbf{86.35}&70.76&92.53&79.24&84.37&83.83\cr
		DGEDT      &86.30&80.00&79.80&75.60&84.00&71.00&91.90&79.00&84.21&83.65\cr
		dotGCN     &86.16&80.49&81.03&78.10&85.24&72.74&93.18&\textbf{82.32}&84.95&\textbf{84.44}\cr
		DualGCN    &87.13&81.16&81.80&78.10&84.50&71.65&91.72&79.46&84.51&84.18\cr \hline
		\rule[0pt]{0pt}{10pt}HyCxG &\textbf{87.32}&\textbf{82.24}&\textbf{82.29}&\textbf{79.11}&86.16&\textbf{74.63}&\textbf{93.83}&82.27&\textbf{85.03}&84.40 \cr
		\bottomrule
	    \end{tabular}}
    \caption{Experimental results on ABSA datasets with BERT encoder. The best result on each dataset is in \textbf{bold}.}
	\label{tab:absa-exp-result}
\end{table*}

%% file: Tables/exp-glue.tex
\begin{table*}[!t]
	\begin{center}
	\fontsize{9}{10}\selectfont
	\resizebox{1.0\linewidth}{!}{
		\begin{tabular}{lcccccccccc}
			\toprule \multirow{2}{*}{\textbf{Model}}&\textbf{CoLA}&\textbf{SST}&\multicolumn{2}{c}{\textbf{MNLI}}&\textbf{QNLI}&\textbf{RTE}&\textbf{QQP}&\textbf{MRPC}&\textbf{STS}& \multirow{2}{*}{\textbf{Avg.}} \\ 
			&Mcc&Acc&Acc&Acc&Acc&Acc&Acc&Acc&Pear    \\ \midrule
			\multicolumn{5}{l}{$\bullet$ \textbf{Results on GLUE development set}}\rule[0pt]{0pt}{10pt}\cr
			RoBERTa$_{\text{B}}$  &63.6&94.8&87.6&87.5&92.8&78.7&91.9&90.2&91.2&86.5 \\ 
			RoBERTa$_{\text{L}}$  &68.0&96.4&90.2&90.2&94.7&86.6&92.2&90.9&92.4&89.1 \\
			HyCxG$_{\text{B}}$    &64.9&95.4&87.8&87.7&93.1&80.9&91.9&90.9&91.7&87.1 \\ 
			HyCxG$_{\text{L}}$    &\textbf{69.6}&\textbf{97.1}&\textbf{90.8}&\textbf{90.4}&\textbf{95.0}&\textbf{89.5}&\textbf{92.3}&\textbf{91.9}&\textbf{92.6}&\textbf{89.9}\\ \midrule
			\multicolumn{10}{l}{$\bullet$ \textbf{Results on GLUE test set} \emph{(from online leaderboard as of Dec., 2022)}}\rule[0pt]{0pt}{10pt}\cr
			RoBERTa$_{\text{B}}$  &60.3&95.7&87.5&87.2&93.0&79.7&89.5&88.1&89.5&85.6 \\ 
			RoBERTa$_{\text{L}}$  &63.0&96.3&89.9&89.4&94.5&85.2&89.6&88.5&91.4&87.5 \\
			HyCxG$_{\text{B}}$    &61.6&96.0&87.7&87.4&93.2&81.2&89.5&88.6&90.7&86.2 \\ 
			HyCxG$_{\text{L}}$    &\textbf{65.9}&\textbf{96.7}&\textbf{90.5}&\textbf{89.9}&\textbf{94.7}&\textbf{86.4}&\textbf{89.7}&\textbf{90.0}&\textbf{91.9}&\textbf{88.4} \\
			\bottomrule
        
		\end{tabular}}
	\end{center}

	\caption{Results for NLU tasks on GLUE development and test set. The best result on each task is in \textbf{bold}. \textbf{Mcc} refers to Matthews correlation coefficient, and \textbf{Pear} refers to Pearson correlation.  \textbf{MNLI} task consists of the \emph{matched} and \emph{mismatched} datasets, while \textbf{SST} and \textbf{STS} denote the SST-2 and STS-B datasets, respectively.}
	\label{tab:experiment_glue}

\end{table*}

%% file: Tables/exp-oml.tex
\begin{table}[t]
    \fontsize{10}{11}\selectfont
	\centering
	\begin{tabular}{p{1.9cm} p{0.9cm}<{\centering} p{0.9cm}<{\centering} p{0.9cm}<{\centering}p{0.9cm}<{\centering}}
		\toprule
	    \textbf{Model}& \textbf{Rest14} & \textbf{Lap14} & \textbf{Rest15} & \textbf{Rest16}  \cr
		\midrule
		OML                 &85.80&80.25&84.32&91.56\cr 
		OML+Pretrain        &86.25&81.35&85.06&92.05\cr  
		OLL+HyCxG           &86.34&80.88&84.69&92.21\cr  \hline
		\textbf{HyCxG}\rule[0pt]{0pt}{10pt}      &\bf 87.32&\bf82.29&\bf86.16&\bf93.83\cr 
	\bottomrule
	\end{tabular}
	\caption{Accuracy comparisons between HyCxG, OML and OLL+HyCxG on ABSA tasks with BERT encoder.}
	\label{tab:experiment-oml}
\end{table}

%% file: Tables/exp-coverage.tex
\begin{table}[t]
    \fontsize{10}{11}\selectfont
	\centering
	\begin{tabular}{p{1.9cm} p{0.9cm}<{\centering} p{0.9cm}<{\centering} p{0.9cm}<{\centering}p{0.9cm}<{\centering}}
		\toprule
	    \textbf{Strategy}& \textbf{Rest14} & \textbf{Lap14} & \textbf{Rest15} & \textbf{Rest16}  \cr
		\midrule
		\textbf{Cond-MC}                     &\bf 87.32&\bf82.29&\bf86.16&\bf93.83 \cr \hline
	    w/o Cond\rule[0pt]{0pt}{10pt}        &85.98&80.41&85.24&92.86\cr 
		w/o Selection                        &86.43&81.03&84.32&92.05\cr  
	\bottomrule
	\end{tabular}
	\caption{Results for different strategies of selecting constructions on four SemEval datasets.}
	\label{tab:experiment-coverage}
\end{table}

%% file: Tables/exp-multilingual.tex
\begin{table}[t]
    \fontsize{10}{11}\selectfont
	\centering
	\begin{tabular}{p{1.7cm} p{0.95cm}<{\centering} p{0.95cm}<{\centering} p{0.95cm}<{\centering} p{0.95cm}<{\centering}}
		\toprule
	    \textbf{Model}& \textbf{French} & \textbf{Spanish} & \textbf{Turkish} & \textbf{Dutch}\cr
		\midrule
		BERT-SPC                 &83.15&90.42&91.72&87.56\cr 
		R-GAT                    &83.84&90.78&92.41&88.32\cr
		KumaGCN                  &83.70&91.79&91.72&88.07\cr
		DGEDT                    &83.43&91.34&93.10&88.32\cr 
		DualGCN                  &83.98&91.24&92.41&88.58\cr \hline
		\textbf{\rule[0pt]{0pt}{11pt}HyCxG} &\bf 84.54&\bf 92.18&\bf94.48&\bf89.34\cr 
	\bottomrule
	\end{tabular}
	\caption{Accuracy comparisons on SemEval multilingual datasets with BERT encoder.}
	\label{tab:experiment-ml}
\end{table}

%% file: Figures/analysis-tsne.tex
\begin{figure}[t]
	\centering
	\includegraphics[width=\columnwidth]{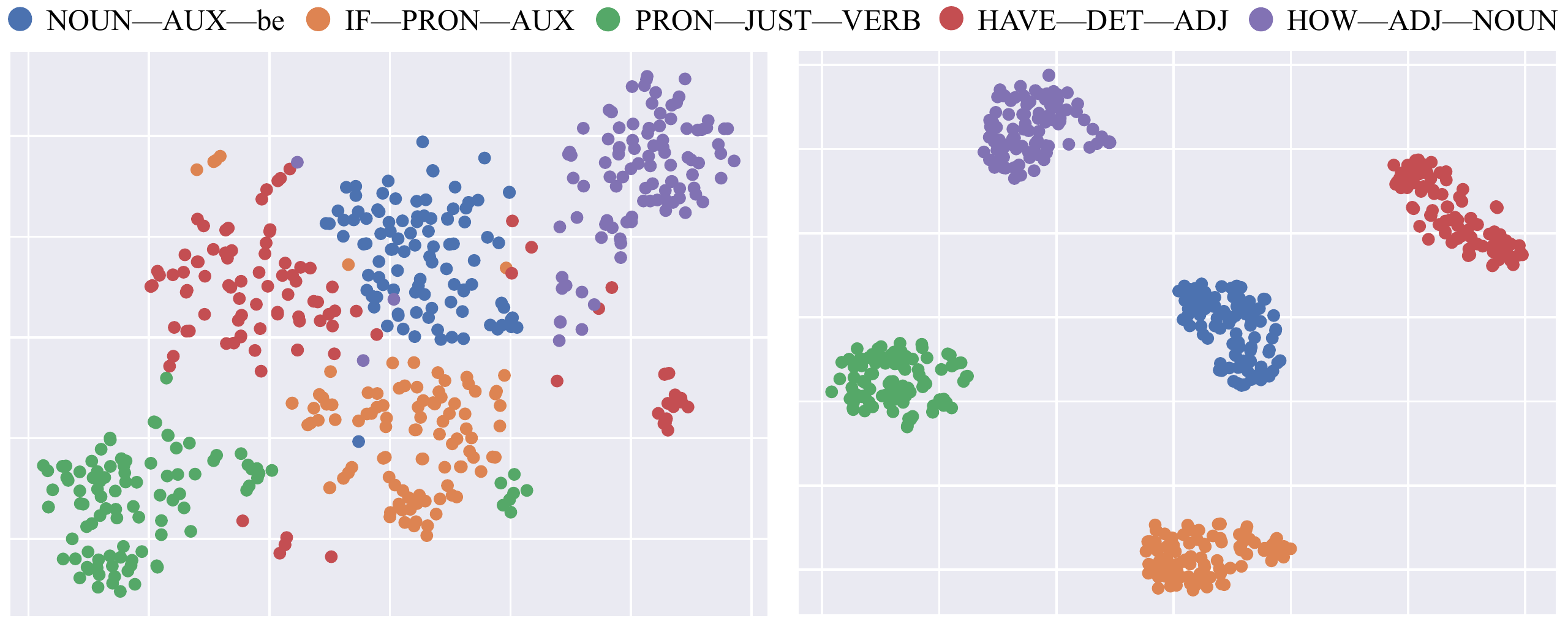} 
	\small (a) BERT \quad \quad \quad \quad \quad \quad \quad \, \, (b) HyCxG
	\caption{2$D$ t-SNE plot of construction representations.}
	\label{fig:analysis-tsne}
\end{figure}

%% file: Figures/analysis-attnvis.tex
\begin{figure}[t]
	\centering
	\includegraphics[width=\columnwidth]{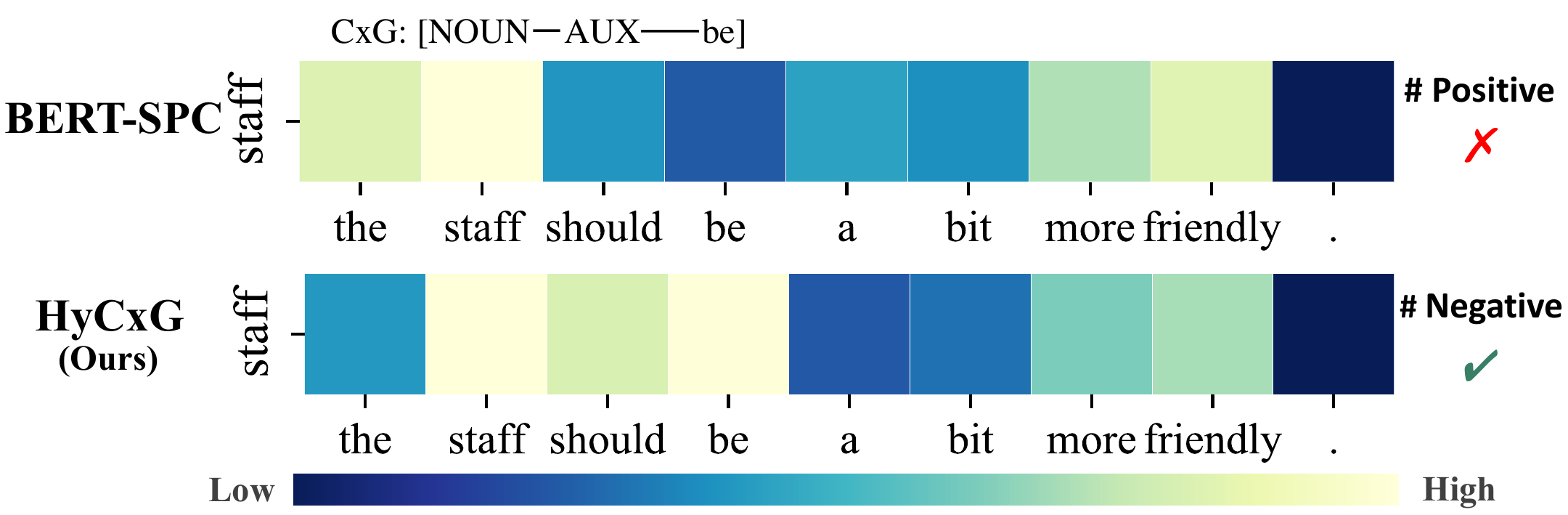} 
	\caption{Visualization of attention scores from BERT-SPC and our HyCxG. Sentiment polarities after the marker \# refer to the prediction label of each model.}
	\label{fig:analysis-attnvis}
\end{figure}

%% file: Chapters/relatedwork.tex
\section{Related Work}
\label{sec:relate-work}

\textbf{Applications of CxG in NLP.} CxG theories have been explored to NLP applications. \citet{doubleday2017processing} exploit embodied CxG to parse the text of requests from users, while \citet{nevens2019computational} employ fluid CxG for semantic parsing in visual question answering. Besides, a knowledge network is generated \cite{icaart20anai} with CxG on winograd schema challenge. These approaches all rely on hand-made definition with prior knowledge and are not compatible with PLMs, thus we employ the usage-based CxG in our work. However, we observe the problems of redundancy and imbalance in usage-based approach, which motivates us to formulate the conditional max coverage problem to acquire discriminative constructions.

\textbf{Probing CxG in PLMs.} Recent work has conducted empirical studies to probe whether PLMs acquire constructions. CxGBERT \cite{tayyar2020cxgbert} finds that constructional information is accessed by BERT but cannot be exploited for lack of proper methods. CxLM \cite{tseng2022cxlm} shows that while PLMs are aware of the construction, they are confused at the variable slots. \citet{weissweiler2022better} discover that PLMs can recognize the comparative correlative constructions but fail to utilize their meaning. Furthermore, \citet{weissweiler2023construction} argue that fine-tuning on downstream tasks necessitates explicit access to constructional information. Our work is the first attempt to exploit the constructional information for representation enhancement on NLU tasks.

\textbf{Hypergraph neural networks.} Recent research has extended conventional graph to hypergraph \cite{feng2019hypergraph,ding2020less,cai2022hypergraph}, which can model high-order data correlation. However, when the networks ignore the representation of hyperedges, the capability of capturing higher-order relationships between nodes can be inhibited \cite{fan2021heterogeneous}. It motivates us to propose R-HGAT to encode constructional information.

\textbf{Aspect-based sentiment analysis.} There has been much work on syntax-based methods \cite{wang2020relational,tang2020dependency,phan2020modelling,li2021dualgraph,tang2022affective,xu2023scope}, which establish syntactic connections between each aspect and their corresponding opinion words via dependency and constituency parsing.

%% file: Chapters/conclusion.tex
\section{Conclusion}
\label{sec:conclusion}

In this work, we introduce usage-based construction grammar (CxG) to enrich language representation with constructional information. Then a hypergraph framework of CxG named HyCxG is proposed to integrate constructional information for language representation enhancement. In HyCxG, we extract constructions through a slot-constraint method, and formulate the conditional max coverage problem for selecting the discriminative constructions. Then we propose a relational hypergraph attention network to encode high-order word interactions within constructions. Extensive experiments illustrate the validity of our HyCxG on natural language understanding tasks.

%% file: Chapters/limitation.tex
\section*{Limitations}
\label{sec:limitation}

In this study, the limitations can be summarized into two major aspects: 

(1) The usage-based approach \cite{dunn2017computational,dunn2019frequency} being employed in our work has the ability to extract most of constructions, while a small portion of non-contiguous constructions (e.g., comparative correlative constructions) are neglected. These non-contiguous constructions are probably fragmented into multiple independent constructions. We will investigate the approaches to capture these non-contiguous constructions via incorporating more syntactic knowledge in future work for language representation enhancement. 

(2) As discussed in Appendix~\ref{sec:colloquial-exp-app}, the performances are not significant improved on the tasks that contain a large amount of colloquial expressions. Since our constructions are mainly learned from the formal corpus, which has less colloquial expressions. It causes fewer constructions to be accessed in these tasks, which encourages us to learn constructions in more diverse corpus to enhance the language representation for natural language understanding tasks in the future. 

%% file: Chapters/acknowledgement.tex
\section*{Acknowledgements}

We  thank all anonymous reviewers for their insightful comments and suggestions. And we also appreciate High-Flyer for providing us with computational cards for pre-training. This research is supported by the Science and Technology Project of Zhejiang Province (2022C01044) and the National Natural Science Foundation of China (51775496).

%% file: Chapters/appendix.tex
\clearpage

\appendix

\section{Background}
\label{sec:appendix-back-ground}
\subsection{Construction Grammar}
\label{sec:appendix-back-cxg}
Construction Grammar (CxG) theory is a branch of cognitive linguistics. It assumes that grammar is a meaningful continuum of lexicon, morphology and syntax rather than solely relying on a system composed of stable but arbitrary rules for generating well-formed sequences. Therefore, constructions (i.e., symbolic elements that connect a certain morphosyntactic form to a meaning) are thought to be the primary objects of grammatical description \cite{langacker1987foundations,goodberg1995constructions,goldberg2006constructions}. Constructions can be defined as linguistic patterns that store different form and meaning pairs. They cannot be strictly predictable by their components. \cite{goodberg1995constructions,langacker2005constructing}. 
Besides, patterns that occur with sufficient frequency can also be considered as constructions \cite{goldberg2006constructions}. 

In terms of the form for CxG, their syntactic structure varies in level of abstractness, including partially or fully filled components (e.g., idioms), and general linguistic patterns \cite{goldberg2003constructions,goldberg2006constructions,dunnwong2022stability}. In another perspective, each construction has immutable components and unfilled slots that the element in the slots can be altered with certain words to boost productivity.

Meanwhile, constructions can also be regarded as linguistic knowledge. This knowledge is based on cognitive results and the product of empirical generalization and abstraction. In the evolution of human language experience, the knowledge is formed, acquired and applied by our human \cite{goldberg2003constructions,hilpert2014construction}.
Analogy to Knowledge Graph (KG), the language systems are considered as a network of constructions \cite{goodberg1995constructions}. This network consists of two fundamental components: (1) nodes in the graph (i.e., specific constructions). (2) the edges between nodes. These edges refer to diverse relationships between constructions (e.g., polysemy link, metaphorical extension and instance link). It is worth mentioning that there may exist multiple connections between constructions in the network.

\subsection{Usage-based Construction Grammar}
\label{sec:usege-based-cxg}
The CxG paradigm has already developed a variety of implementations, including formal approaches, i.e., fluid construction grammar (FCG), embodied construction grammar (ECG) and sign-based construction grammar (SBCG) as well as usage-based approaches. However, the FCG, ECG and SBCG rely heavily on hand-made definition with prior knowledge \cite{steels2006brief,goldberg2005construction,rambelli2019distributional}. Though these methods can provide high-quality representations, they cannot automatically mine the constructions from the data with the emergence of slot-constraints \cite{dunn2017computational,dunn2019frequency}.

We follow the efforts of \citet{dunn2019frequency} that propose a computationally slot-constraints approach for CxG extraction via the data-driven pipeline. First, three types of slots (i.e., lexical, syntactic and joint semantic-syntactic) are defined according to the different levels of abstraction. Note that the syntactic slots refer to the POS tags, while the semantic-syntactic slots are derived from word embeddings that are clustered in discrete semantic domains based on K-Means algorithm \cite{dunnwong2022stability}.
Secondly, they generate the potential construction templates from the corpus. Then the statistics of frequency and association strength are calculated for pruning templates.
Finally, the tabu search is applied to determine the optimal set of constructions with Minimum Description Length (MDL) metric \cite{dunn2019frequency}. Based on this pipeline, they develop the c2xg\footnote{\url{https://pypi.org/project/c2xg/}} toolkit.

\input{Tables/stat-glue}

\section{Data Statistics}
\label{appendix:data-statistic}

\input{Tables/stat-absa}

We conduct experiments on a variety of natural language understanding tasks in main experiments, including eight tasks from the GLUE benchmark \cite{wang2018glue} and five aspect-level sentiment analysis tasks \cite{pontiki2014semeval,pontiki2015semeval,pontiki2016semeval,jiang2019challenge}. 
Table~\ref{tab:statistic-glue} presents the data statistics for GLUE benchmark tasks. We divide the GLUE benchmark tasks into two categories: single-sentence task and sentence-pair task. In single-sentence task, the language acceptability task (CoLA) and the sentence-level sentiment task (SST-2) are included. As for sentence-pair task, it contains similarity task (STS-B), paraphrasing tasks (MRPC, QQP) and several natural language inference (NLI) tasks (MNLI, QNLI and RTE). All tasks except STS-B are classification tasks, while STS-B is a regression task.

The statistics for aspect-based sentiment analysis datasets are shown in Table~\ref{tab:statistic-absa}. They are all annotated with aspects and corresponding polarities (i.e., positive, neutral or negative). In multilingual settings, we adopt four different language datasets from SemEval 2016 \cite{ pontiki2016semeval}. Meanwhile, Twitter \cite{dong2014adaptive} and GermEval \cite{wojatzki2017germeval} datasets, which collect tweets and messages from social media, are employed in the colloquial expression experiments.

\section{Learning and Alignment of CxG}
\label{sec:appendix-cxg-learn}
In order to utilize constructional information, the inventory of construction grammar needs to be learned at first. As we present a brief description of construction learning procedure in Section~\ref{sec:usege-based-cxg}, prior work has shown that construction grammars can converge on stable representations with sufficient training data \cite{dunn2022exposure}. Therefore, the construction grammar in English is learned from sampling of multi-source corpus (\textsc{Wikipedia}, \textsc{BookCorpus} and \textsc{CC}-\textsc{NEWS}) \cite{devlin2019bert,liu2019roberta} which contains about 1,200 million words. As for other languages utilized in multilingual experiments, constructions are learned from sampling the multilingual portion of the \textsc{C4} corpus \cite{raffel2020exploring}. Since the constructions on a variety of languages are learned in c2xg toolkit, we validate that our HyCxG can achieve comparable performances on NLU tasks with the construction grammar list in c2xg.

After obtaining the construction grammar list, we are able to derive all possible constructions from a given sentence via c2xg toolkit. However, the toolkit can only detect which construction patterns are present in the sentences and cannot provide specific position indexes of them. Besides, there are discrepancies between c2xg and pre-trained language models in tokenization procedure. The c2xg toolkit utilizes the basic white-space tokenizer, while pre-trained language models adopt more complex algorithms (e.g., WordPiece \cite{devlin2019bert}, SentencePiece \cite{kudo2018sentencepiece}). Therefore, we have to align the position indexes of the constructions under different tokenization algorithms with specific mapping function. 

To tackle these imperatives, we develop a wrapper for c2xg. It can be adapted to different tokenization methods with the output of the start and end position index for construction spans. We hope that it can facilitate future research on construction grammar in natural language processing tasks.

\section{Analysis of Computational Complexity}
\label{sec:appendix-complexity}

\input{Tables/exp-complexity}
\label{sec:computation-complexity-app}
In Table~\ref{tab:experiment-oml}, we demonstrate that the effectiveness of our model does not result from the increased complexity of the model. As shown in Table~\ref{tab:appendix-complexity}, we further analyze the computational complexity of the models via \textsc{\href{https://www.deepspeed.ai/tutorials/flops-profiler}{DeepSpeed}} to illustrate more intuitive results. The number of model parameters and the multiply–accumulate operations (MACs) are utilized to compare complexity on aspect-based sentiment analysis models. To ensure a fair comparison, we unify the public hyper-parameters of all baselines while preserving other unique hyper-parameters consistent with their official implementations. Besides, the additional embedding parameters for constructions in HyCxG are not counted. Then three observations can be derived. 

First, LCFS-ASC is significantly more complex than other models due to its dual BERT architecture, while other models all use a single PLM. Second, adding an additional transformer layer to BERT (OML) indeed improves performances but is far more complex than our HyCxG. However, our model outperforms OML, which proves the improvement of our HyCxG is not just due to the increased complexity of the model structure. Third, our model has lower computational cost and memory consumption than most of baselines. Meanwhile, our HyCxG achieves even better performances without dependency information injection, which empirically validates the effectiveness of constructional information. 

\section{Hyper-parameters and Settings}
\label{sec:appendix-hyper-params}

For fine-tuning our HyCxG on GLUE and aspect-based sentiment analysis datasets, we search for the optimal task-specific hyper-parameters with the range of values in Table~\ref{tab:appendix-hyperfinetune} (We implement our model in \textsc{Pytorch} and use GeForce RTX 3090 devices for experiments):

\input{Tables/exp-hyperfine}
\input{Tables/exp-counterfactual}
In order to pre-train OML models in Table~\ref{tab:experiment-oml} and~\ref{tab:experiment-oml-glue}, we follow the same pre-training procedure of BERT and RoBERTa. OML models are implemented in \textsc{Pytorch} and pre-trained on 8 $\times$ Tesla A100 devices. We employ the officially released PLM checkpoints from Hugging Face\footnote{\url{https://huggingface.co/models}} to initialize basic parameters (12 layers) in OML, while the parameters in additional transformer layer are initialized to the average of the preceding 12 layers. Besides, we set the batch size to 64 per GPU device, and the training steps to 400,000 during pre-training procedure with AdamW optimizer (learning rate is 5e-5 and warmup ratio is 0.1) and the corpus of \textsc{Wikipedia} and \textsc{BookCorpus}.

\section{Pattern Recognition Capability of CxG}
\label{sec:counterfactual-pattern-app}
In this work, we demonstrate that the construction grammar is capable of improving natural language understanding via infusing constructional information. However, can CxG only serve to enhance language representation? We explore more application scenarios for construction grammar based on this concern and a viable scene is pattern recognition. The identification of this scenario can be viewed as an investigation into whether the text contains particular specific structures (e.g., negation and counterfactual expression). We take counterfactual detection task as an example to verify the feasibility of the scenario. 

The counterfactual dataset in SemEval 2020 \cite{yang2020semeval} are adopted in this experiment with train (13,000 instances) and test (7,000 instances) sets. We employ Precision, Recall and F$_1$ score as the metrics to evaluate the performance. To better evaluate our method, we compare the performances with the best system in SemEval. HIT \cite{ding2020hit} is the state-of-the-art method that employs the ensemble model (combining the large version of BERT, RoBERTa and XLNet via weighted average for their probability predictions) with pseudo-labeling \cite{lee2013pseudo} strategy and 10-fold cross-validation. Table~\ref{tab:experiment-counterfactual} shows the results and the above system is marked as HIT+PL. HIT is the single model of RoBERTa-large without pseudo-labeling. Our HyCxG model is also trained with 10-fold cross-validation based on RoBERTa-large, while HyCxG+PL incorporates pseudo-labeling strategy to HyCxG.

Surprisingly, we observe that our HyCxG significantly outperforms HIT system by 4.47 F$_1$ points. Even for HIT+PL (applying ensemble and pseudo-labeling), there is still a massive gap with HyCxG. Furthermore, our HyCxG+PL also has a 3.82 F$_1$ points boost compared to HIT+PL. It endorses the ability of our model in pattern recognition.

As discussed by \citet{yang2020semeval}, the main challenge in counterfactual detection task is that existing models focus excessively on token level features while neglecting to understand statements. Most counterfactual cases are expressed with subjunctive mood to convey wishes, suggestions and demands. There is a practical example ``\emph{if I were asked to, I would be happy to talk to anyone.}'', which is misclassified by HIT model, while HyCxG can predict correctly. It is a counterfactual expression that regards the part guided by the conjunction ``\emph{if}'' as the antecedent. Based on such hypothesis, the possible consequent is stated. In this sentence, our system can extract two critical construction ``\emph{if}--PRON--\emph{were}--VERB'' $\to$ ``\emph{if I were asked}\:\!'' and ``AUX--\emph{be}--ADJ'' $\to$ ``\emph{would be happy}'' for capturing the counterfactual statements. Thus, HyCxG is capable of recognizing specific patterns via encoding constructional information.

Similarly, negation detection is also a recognition task relying on a series of special patterns. There are more application scenarios of CxG, which we leave those discussions to future work.

\input{Tables/exp-oml-glue}
\section{Ablation Study on GLUE Datasets}
\label{sec:ablation-on-glue-app}
Similar to the ablation study on ABSA tasks (Table~\ref{tab:experiment-oml}), we conduct experiments on GLUE benchmark with four datasets (CoLA, SST-2, RTE and MRPC). As discussed previously, the PLM (here is RoBERTa-base) with an additional layer on top is called OML. OML+Pretrain performs an extra pre-training procedure on OML (details of pre-training are shown in Section~\ref{sec:appendix-hyper-params}). OLL+HyCxG is the model that removes the last transformer layer from PLM in our HyCxG model. For the results in Table~\ref{tab:experiment-oml-glue}, OML+Pretrain always outperforms OML on four tasks, which illustrates the utility of pre-training procedure. Besides, our OLL+HyCxG achieve better performances compared to OML, even though it has two fewer transformer layers. It is worth mentioning that the result of OLL+HyCxG is even higher than OML+Pretrain on SST-2 dataset.

Moreover, HyCxG performs better than OML and OML+Pretrain, while it has a smaller number of parameters with lower complexity. These results indicate the benefits of constructional information in natural language understanding tasks and high cost performance of our HyCxG.

\section{Colloquial Expression Results}
\label{sec:colloquial-exp-app}

\input{Figures/appendix-informal}

As we discuss in \nameref{sec:limitation}, the performance improvement of HyCxG is not significant for datasets with informalized expressions (i.e., colloquial expressions). In this work, constructions are learned from the standard corpus (e.g., \textsc{Wikipedia} and \textsc{BookCorpus}) with formal language. This causes some of edge constructions not being extracted for language representation enhancement. We conduct experiments on Twitter \cite{dong2014adaptive} and GermEval \cite{wojatzki2017germeval} datasets to investigate 
the issue for English and German. These datasets are collected from social media which consist of numerous colloquial sentences (statistics are shown in Table~\ref{tab:statistic-absa}). As shown in Table~\ref{tab:experiment-informal}, the results on Twitter are retrieved from the papers of baselines. Then we adapt all the models with multilingual settings (Section~\ref{sec:multilingual}) to the GermEval dataset. From the results, we observe the performance of our HyCxG on Twitter is only in the middle compared to baselines. As for GermEval, HyCxG exceeds most of models but also falls below DualGCN.

\input{Tables/exp-informal}
\label{sec:appendix-informal}
To further analyze the evidence for supporting our conclusion, two metrics are proposed to evaluate the sparsity of constructions: average ratio of construction (AoC) and average coverage ratio (ACR). We define AoC as the average proportion of extracted constructions in each sentence, while ACR means the average ratio for the total length of the constructions to sequence length in each sentence under maximum coverage (Section~\ref{sec:max-coverage}). Metrics can also be formulated as follows:
\begin{equation}
\label{eq:informal-metric}
    \begin{aligned}
        \operatorname{AoC} &= \frac{1}{N}\sum_{i=1}^N \frac{ \operatorname{CxG}\left(s_i\right)}{\mathcal{L}\left(s_i\right)} \\
        \operatorname{ACR} = \frac{1}{N} &\sum_{i=1}^N \frac{\sum_{j=1}^M \mathcal{L}\left(c_j\right)}{\mathcal{L}\left(s_i\right)}, c_j\in \operatorname{MC\left(s_i\right)}
    \end{aligned}
\end{equation}
where $N$ is the number of sentences in each dataset while $s_i$ is the $i$-th sentence. $\mathcal{L}(\cdot)$ is the function, that computes the length of the input and $\operatorname{CxG}(\cdot)$ is employed to calculate the total number of constructions in a sentence. Besides, $\operatorname{MC}(\cdot)$ refers to the set of constructions in the sentence that satisfy the maximum coverage (w/o condition), while $M$ is the size of the set and $c_j$ is the $j$-th construction.

We compare these metrics between formal group (Rest14, Lap14) and colloquial group (Twitter, GermEval) and the results are shown in Figure~\ref{fig:appendix-informal}. There is a huge gap, where colloquial group is significantly lower than formal group on AoC and ACR. It demonstrates that constructions are much sparser in Twitter and GermEval. In such circumstances, constructional information learned by our model has been inadequate so far to enhance the representation. For this limitation in our work, we believe that it can be improved via adding more colloquial corpus to the construction learning (Section~\ref{sec:appendix-cxg-learn}), which requires further investigations in the future. 

\section{Ablation Study on Model Components}
\label{sec:appendix-component}
\input{Tables/exp-ablation}

To further investigate the influence of different components in our HyCxG, we conduct extensive ablation studies. As shown in Table~\ref{tab:experiment-ablation-component}, w/o HGATT and FFN  indicate that we temporarily remove each of these components from our model. The experimental results illustrate the importance of the HGATT network, since its removal can cause significant performance degradation. Meanwhile, it also demonstrates the validity of injecting constructional information for language representation enhancement. Overall, our HyCxG with all components can achieve the highest performance.

\section{Network of Constructions}
\label{sec:constructicon-app}

Construction grammar is not an unordered set in a language, but rather a network linked by inheritance relations. The whole network of constructions in a particular language is called the \emph{Constructicon} \cite{evans2007glossary}. As briefly introduced in Section~\ref{sec:appendix-back-cxg}, \emph{Constructicon} consists of nodes (constructions) and edges (inheritance relations) \cite{goldberg2003constructions}. There are four types of relations between constructions: polysemy link ($\operatorname{I}_P$), subpart link ($\operatorname{I}_S$), instance link ($\operatorname{I}_I$) and metaphorical extension ($\operatorname{I}_M$). Previous efforts are taken by linguists using case studies to manually build the networks \cite{boogaart2014extending}. To the best of our knowledge, there is no attempt to construct \emph{Constructicon} from the perspective of distributional representation. In this work, we can acquire representation of each construction via the embedding matrix in HyCxG. Therefore, our HyCxG can  contribute to linguistics in turn to provide interpretability for the network of constructions.

\input{Algorithms/algo-med}

We discuss three types of inheritance relationships ($\operatorname{I}_P$, $\operatorname{I}_S$ and $\operatorname{I}_I$) in this work, while metaphorical extension is more abstract and complex. $\operatorname{I}_P$ characterizes the semantic relationship between a specific meaning of a construction and its extended meaning. $\operatorname{I}_S$ is referred to the connection that a construction is an independent subpart of another construction. Besides, when a concrete construction is a special instance of another construction, the relationship is defined as $\operatorname{I}_I$. For the purpose of modeling inheritance relationships, we measure the distances between different constructions from semantic and morphological perspectives. Then we weight the sum of the two distances and the top-k closest constructions are selected for each construction to determine the relationships. Eventually, the network of constructions is generated. 

\input{Tables/exp-constructicon}

Since the representation of each construction is obtained, semantic distance (SD) can be computed via cosine similarity. Meanwhile, we propose multi-level edit distance (MED) algorithm to evaluate the morphological distance (MD) between two constructions with different abstract level. In this setting, semantic distance can provide evidence for polysemy link, while morphological distance constrains variation in structures for subpart link and instance link. Thus, these distances are complementary on modeling inheritance relations. Under definition, semantic distance can be formulated as:
\begin{equation}
   \operatorname{SD}\left(g_i, g_j\right) = 1 - \frac{\overrightarrow{v_{g_i}} \cdot \overrightarrow{v_{g_j}}}{\Vert \overrightarrow{v_{g_j}}\Vert \Vert\overrightarrow{v_{g_j}} \Vert}
\end{equation}
where $\overrightarrow{v_{g_i}}$ and $\overrightarrow{v_{g_i}}$ are the embedding vector of construction $g_i$ and $g_j$, which can be looked up from construction embedding matrix $E_{c}$, respectively.

MED is proposed to compute the morphological distances. As slots in constructions are defined for different levels (Lexical, Syntactic and Semantic) of abstraction, we need to match slots across the levels. We define a function \texttt{Is\_Match}($\cdot, \cdot$) to indicate whether two slots match or not. The matching criteria can be summarized as follows:
\begin{itemize}[itemsep=3pt,topsep=1pt,parsep=0pt]
    \item (Lexical$\leftrightarrow$Lexical) Match when the slots are identical with the same POS tags or belong to common semantic cluster.
    \item (Syntactic$\leftrightarrow$Syntactic, Semantic$\leftrightarrow$Semantic) Match if and only if the slots are identical.
    \item (Lexical$\leftrightarrow$Syntactic) Match when the slot of lexical has the same POS tags as the other.
    \item (Lexical$\leftrightarrow$Semantic) Match when the semantic cluster of the lexical slot is identical to the slot in semantic level.
\end{itemize}

Based on such matching strategy, MED algorithm is illustrated in Algorithm~\ref{algo:med}. After obtaining semantic and morphological distances, we synthesize them to compute the construction distance. Then the top-k closest constructions for each construction can be selected as the potential inheritance set. It can be derived as:
\begin{equation}
   \{\mathcal{G}_g \subset \mathcal{G} \vert \underset {\vert \mathcal{G}_g \vert = k} {\operatorname {arg\,min}}(\gamma \operatorname{SD}_g + \operatorname{MD}_g)\}
   \label{eq:constructicon-topk}
\end{equation}
where $ \mathcal{G}_g$ is the potential inheritance set for construction $g$, while $\mathcal{G}$ is the set of all constructions. $\operatorname{SD}_g$ and $\operatorname{MD}_g$ refer to semantic and morphological distances between $g$ and the other constructions in $\mathcal{G}$, respectively. $\gamma$ is the coupling co-efficiency that regulates the two distances. Then we can determine the inheritance relationship within the set based on the definition of $\operatorname{I}_P$, $\operatorname{I}_S$ and $\operatorname{I}_I$.

\input{Figures/appendix-constructicon}

In order to analyze \emph{Constructicon}, the embedding matrix $E_c$ of constructions is extracted after training on CoLA task as an example. We set $k$ to 15 and $\gamma$ to 10 in Equation~\ref{eq:constructicon-topk}. In such settings, we are able to illustrate the effectiveness of our approach with practical examples for each inheritance relationship. As shown in Table~\ref{tab:appendix-constructicon}, the relationship between ``\emph{think}--PRON--AUX'', ``\emph{believe}--PRON-AUX'' and ``\emph{know}--PRON--AUX'' belongs to polysemous inheritance. They can all be considered as instantiations of construction ``VERB--PRON--AUX''. Thus, the subjective belief that someone has done or can do something is the primary meaning of this type of constructions. After that, different verbs are filled in the slots to inject expanded meanings. As for subpart inheritance, ``SCONJ--PRON--AUX'' is the smallest independent construction in the example. It is embodied in ``SCONJ--PRON--AUX--VERB'',  while ``SCONJ--PRON--AUX--VERB'' is also a subpart of ``SCONJ--PRON--AUX--VERB--\emph{to}''. The instance inheritance establishes the connection between the construction and its completely or partially filled constructions . ``\emph{if}--PRON--\emph{can}'' is a specific construction that fills ``\emph{if}'' and ``\emph{can}'' into ``SCONJ--PRON--AUX'' as a substitute for ``SCONJ'' and ``AUX''. Besides, it is worth mentioning that $\operatorname{I}_P$ is undirected relationship, while $\operatorname{I}_S$ and $\operatorname{I}_I$ are directed relationships in our \emph{Constructicon}.

Furthermore, we develop a visualization toolkit to demonstrate the network of constructions. As shown in Figure~\ref{fig:appendix-constructicon}, a sub-network that depicts the three inheritance relationships is constructed as an example under our settings.  

\section{Detailed Results on ABSA Tasks}

\input{Tables/exp-validation-absa}

Since all four ABSA datasets (Rest14, Lap14, Rest15 and Rest16) of SemEval do not have official validation sets, randomness may be introduced for performance comparison. Therefore, we randomly divide the training set into a new training set and a validation set with the ratio of 9:1 for these ABSA datasets. Then we conduct the experiments with baseline models and our HyCxG. Meanwhile, the hyper-parameter searching is conducted for each model for fair comparison.

As shown in Table~\ref{tab:experiment-validation-absa}, we can observe that the performance comparison of the model is almost consistent with that in Table~\ref{tab:absa-exp-result}. First, the baseline models that inject syntactic information always achieve better performances than BERT-SPC. Second, R-GAT outperforms the other baseline models on the Rest14 and Rest15 datasets, while KumaGCN and DualGCN have better performance on Lap14 and Rest16, respectively. Third, HyCxG outperforms all baseline models with constructional information incorporated on these datasets, which illustrates the validity of HyCxG. Meanwhile, it can also substantiate that construction grammar can be regarded as the inductive bias to enhance the language representation in NLU tasks.

%% file: Tables/stat-glue.tex
\begin{table}[t]	
	\centering
	\fontsize{10}{12}\selectfont
	\begin{tabular}{p{1.0cm}<{\centering} p{0.6cm}<{\centering} p{1cm}<{\centering} p{1cm}<{\centering} p{2cm}<{\centering}}
    \toprule
    \textbf{Dataset} & \textbf{$|L|$} & \textbf{\#Train} & \textbf{\#Test} & \textbf{\#Task}\\ \midrule
    \multicolumn{5}{l}{$\bullet$ \textbf{Single-sentence task}}\rule[0pt]{0pt}{10pt}\cr
    \textbf{CoLA}  & 2 & 8,551 & 1,063 & Acceptability \cr
    \textbf{SST-2} & 2 & 67,349 & 1,821 & Sentiment \cr \hline
    \multicolumn{5}{l}{$\bullet$ \textbf{Sentence-pair task}}\rule[0pt]{0pt}{10pt}\cr
    \textbf{QQP}    & 2 & 363,846 & 390,965 & Paraphrase \cr
    \textbf{MRPC}   & 2 & 3,668 & 1,725 & Paraphrase \cr
    \textbf{MNLI}   & 3 & 392,702 & 19,643 & NLI \cr
    \textbf{QNLI}   & 2 & 104,743 & 5,463 & NLI \cr
    \textbf{RTE}    & 2 & 2,490 & 3,000 & NLI \cr
    \textbf{STS-B}  & * & 5,749 & 1,379 & Similarity \cr
    \bottomrule 
    \end{tabular}
	\caption{Statistics for the GLUE benchmark. $|L|$ indicates the number of classes for classification tasks, while * refers to regression task. The labels of STS-B are continuous values from 0 to 5.}
	\label{tab:statistic-glue}
\end{table}

%% file: Tables/stat-absa.tex
\begin{table}[t]	
	\centering
	\fontsize{10}{12}\selectfont
	\begin{tabular}{p{1.3cm}<{\centering} p{1.0cm}<{\centering} p{1.1cm}<{\centering} p{1.1cm}<{\centering} p{1.1cm}<{\centering}}
    \toprule
    \textbf{Dataset} & \textbf{Division} & \textbf{\#Pos.} & \textbf{\#Neu.} & \textbf{\#Neg.}\\ \midrule
    \multicolumn{5}{l}{$\bullet$ \textbf{Main Experiments}}\rule[0pt]{0pt}{10pt}\cr
    \multirow{3}{*}{\rule[0pt]{0pt}{10pt}\textbf{MAMS}} & Train & 3,380 & 5,042 & 2,764 \cr
    & Develop & 403 & 604 & 325 \cr
    & Test & 400 & 607 & 329 \cr \hline
    \multirow{2}{*}{\textbf{Rest 14}}& Train & 2,164 & 637 & 807 \cr
    & Test & 728 & 196 & 196 \cr \hline
    \multirow{2}{*}{\textbf{Lap 14}}& Train & 994 & 464 & 870 \cr
    & Test & 341 & 169 & 128\cr \hline
    \multirow{2}{*}{\textbf{Rest 15}} & Train & 912 & 36 & 256 \cr
    & Test & 326 & 34 & 182 \cr \hline
    \multirow{2}{*}{\rule[0pt]{0pt}{10pt}\textbf{Rest 16}} & Train & 1,240 & 69 & 439 \cr
    & Test & 469 & 30 & 117 \cr \hline \specialrule{0em}{0.8pt}{0.8pt} \hline
    \multicolumn{5}{l}{$\bullet$ \textbf{Multilingual Experiments}}\rule[0pt]{0pt}{10pt}\cr
    \multirow{2}{*}{\rule[0pt]{0pt}{10pt}\textbf{French}} & Train & 901 & 116 & 753 \cr
    & Test & 364 & 69 & 285 \cr \hline
    \multirow{2}{*}{\rule[0pt]{0pt}{10pt}\textbf{Spanish}} & Train & 1,368 & 89 & 479 \cr 
    & Test & 521 & 34 & 176 \cr \hline
    \multirow{2}{*}{\rule[0pt]{0pt}{10pt}\textbf{Turkish}} & Train & 770 & 111 & 504 \cr 
    & Test & 93 & 6 & 46 \cr \hline
    \multirow{2}{*}{\rule[0pt]{0pt}{10pt}\textbf{Dutch}} & Train & 758 & 118 & 407 \cr
    & Test & 245 & 24 & 125 \cr \hline \specialrule{0em}{0.8pt}{0.8pt} \hline
    \multicolumn{5}{l}{$\bullet$ \textbf{Colloquial Experiments}}\rule[0pt]{0pt}{10pt}\cr
    \multirow{2}{*}{\rule[0pt]{0pt}{10pt}\textbf{Twitter}} & Train & 1,561 & 3,127 & 1,560 \cr
    & Test & 173 & 346 & 173 \cr \hline
    \multirow{3}{*}{\rule[0pt]{0pt}{10pt}\textbf{GermEval}} & Train & 1,592 & 812 & 4,554 \cr
    & Develop & 207 & 83 & 777 \cr
    & Test & 127 & 175 & 1,112 \cr 
    \bottomrule 
    \end{tabular}
	\caption{Statistics for ABSA datasets that be employed in the main, multilingual and colloquial experiments.}
	\label{tab:statistic-absa}
\end{table}

%% file: Tables/exp-complexity.tex
\begin{table}[t]
    \fontsize{10}{11}\selectfont
	\centering
	\begin{tabular}{p{2.0cm} p{1.3cm}<{\centering} p{1.4cm}<{\centering} p{1.3cm}<{\centering}}
		\toprule
	    \textbf{Model} & \textbf{Params} & \textbf{w/o PLM} & \textbf{MACs}   \cr
		\midrule
		BERT-SPC           &109.49M&0.01M&13.16G\cr 
		R-GAT              &110.45M&0.97M&13.20G\cr 
		LCFS-ASC           &\textbf{228.42}M&\textbf{9.46}M&\textbf{26.79}G\cr 
		KumaGCN            &112.49M&3.01M&13.87G\cr 
		DGEDT              &112.50M&3.02M&14.08G\cr 
		DualGCN            &111.85M&2.37M&13.74G\cr \hline
		\rule[0pt]{0pt}{11pt}OML &116.57M&7.09M&14.25G\cr 
		HyCxG              &112.24M&2.76M&13.48G\cr 
	\bottomrule
	\end{tabular}
	\caption{Computational complexity analysis of models. \textbf{Params}, \textbf{w/o PLM} and \textbf{MACs} represent the number of parameters, model parameters without PLM and multiply–accumulate operations, respectively.}
	\label{tab:appendix-complexity}
\end{table}

%% file: Tables/exp-hyperfine.tex
\begin{table}[ht]
\fontsize{10}{12}\selectfont
\begin{tabular}{p{2.8cm} p{4.0cm}}
\toprule
\textbf{Configuration} & \textbf{Value} \\ \midrule
Epoch                     & 4 / 10 / 50\cr 
Batch size                & 16 / 24 / 32 \cr
Sequence length           & 150 / 250 / 300\cr
Learning rate             & 1e-5 / 2e-5 / 3e-5 \cr
Dropout (PLM)             & [0.0, 0.5]\cr
Dropout (R-HGAT)          & [0.0, 0.5] \cr
Layers                    & 1 \cr
Weight decay              & 1e-1 / 1e-2 \cr
Warmup ratio              & 0.03 / 0.06 \cr
\bottomrule 
\end{tabular}

\caption{Detailed hyper-parameter configurations.}

\label{tab:appendix-hyperfinetune}
\end{table}

%% file: Tables/exp-counterfactual.tex
\begin{table}[t]
    \fontsize{10}{11}\selectfont
	\centering
	\begin{tabular}{p{1.3cm} p{1.2cm}<{\centering} p{1.1cm}<{\centering} p{1.1cm}<{\centering} p{0.8cm}<{\centering}}
		\toprule
	    \textbf{Models}& \textbf{Ensemble} & \textbf{Precision} & \textbf{Recall} & \textbf{F$_1$}   \cr
		\midrule
		HIT                                 &\usym{2717}&88.27&90.76&89.49 \cr 
		HIT+PL                              &\usym{2713}&91.90&90.00&90.90 \cr \hline
	    \rule[0pt]{0pt}{11pt}HyCxG          &\usym{2717}&94.09&93.83&93.96 \cr 
		HyCxG+PL                            &\usym{2717}& \bf 95.21& \bf94.24& \bf94.72 \cr  
	\bottomrule
	\end{tabular}
	\caption{Results on counterfactual detection dataset. \textbf{PL} is the abbreviation for pseudo-labeling.}
	\label{tab:experiment-counterfactual}
\end{table}

%% file: Tables/exp-oml-glue.tex
\begin{table}[t]
    \fontsize{10}{11}\selectfont
	\centering
	\begin{tabular}{p{1.9cm} p{0.9cm}<{\centering} p{0.9cm}<{\centering} p{0.9cm}<{\centering}p{1.0cm}<{\centering}}
		\toprule
	    \textbf{Model}& \textbf{CoLA} & \textbf{SST-2} & \textbf{RTE} & \textbf{MRPC}  \cr
		\midrule
		OML                 &64.24&95.07&79.06&90.44\cr 
		OML+Pretrain        &64.84&95.18&80.51&91.18\cr  
		OLL+HyCxG           &64.42&95.30&79.78&90.69\cr  \hline
		\textbf{HyCxG}\rule[0pt]{0pt}{11pt}  &\bf 64.90&\bf95.41&\bf80.87&\bf90.93\cr 
	\bottomrule
	\end{tabular}
	\caption{Ablation study on four GLUE development datasets with RoBERTa-base encoder.}
	\label{tab:experiment-oml-glue}
\end{table}

%% file: Figures/appendix-informal.tex
\begin{figure}[t]
	\centering
	\includegraphics[width=\columnwidth]{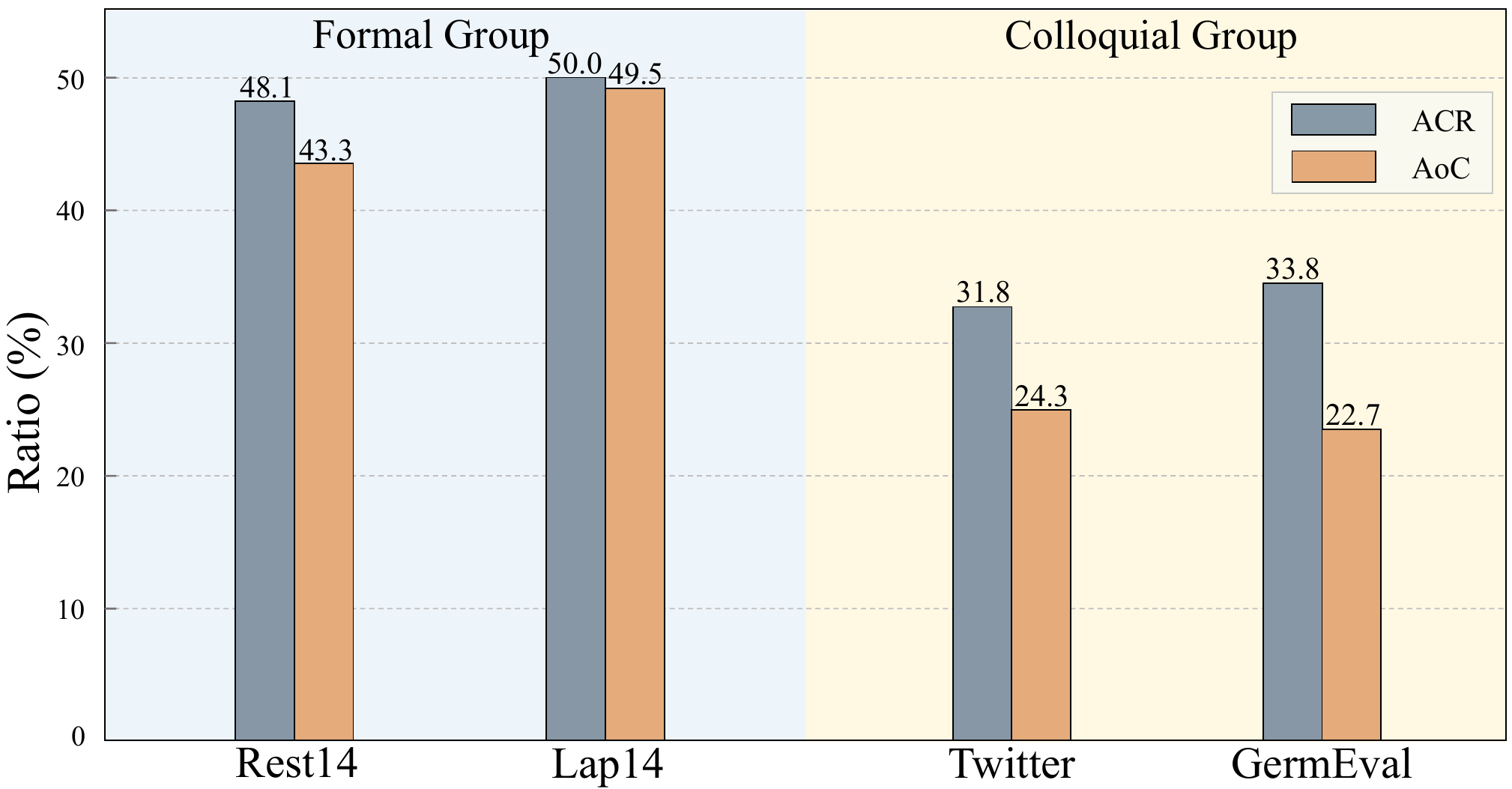} 
	\caption{Comparison of construction sparsity based on AoC and ACR metrics.}
	\label{fig:appendix-informal}
\end{figure}

%% file: Tables/exp-informal.tex
\begin{table}[t]
    \fontsize{10}{11}\selectfont
	\centering
	\begin{tabular}{p{1.7cm} p{0.95cm}<{\centering} p{0.95cm}<{\centering} p{0.95cm}<{\centering} p{0.95cm}<{\centering}}
		\toprule
	    \multirow{2}{*}{\rule[0pt]{0pt}{12pt}\textbf{Model}}
	     & \multicolumn{2}{c}{\textbf{Twitter}} 
	     & \multicolumn{2}{c}{\textbf{GermEval}} \cr \cline{2-5}\rule[0pt]{0pt}{11pt}
	     &\textbf{Acc}&\textbf{F$_1$}&\textbf{Acc}&\textbf{F$_1$} \cr
		\midrule
		BERT-SPC                 &75.14&73.59&84.65&66.76\cr 
		R-GAT                    &76.15&74.88&85.50&68.81\cr 
	    KumaGCN                  &77.89&\textbf{77.03}&85.08&69.26\cr
		DGEDT                    &\textbf{77.90}&75.40&86.71&72.81\cr 
		DualGCN                  &77.40&76.02&\bf 86.96&\bf 73.44\cr \hline
		\textbf{\rule[0pt]{0pt}{11pt}HyCxG} & 77.17& 75.70& 86.85&71.71\cr 
	\bottomrule
	\end{tabular}
	\caption{Experimental performances of datasets with colloquial expressions.}
	\label{tab:experiment-informal}
\end{table}

%% file: Tables/exp-ablation.tex
\begin{table}[t]
    \fontsize{10}{11}\selectfont
	\centering
	\begin{tabular}{p{2.25cm} p{0.83cm}<{\centering} p{0.83cm}<{\centering} p{0.83cm}<{\centering}p{0.83cm}<{\centering}}
		\toprule
	    \textbf{Model}& \textbf{Rest14} & \textbf{Lap14} & \textbf{Rest15} & \textbf{Rest16}  \cr
		\midrule
		HyCxG                                    &\bf 87.32&\bf82.29&\bf86.16&\bf93.83\cr \hline
		w/o HGATT\rule[0pt]{0pt}{10pt}           &86.34&81.19&84.69&92.37\cr
		w/o FFN                                  &86.70&81.50&85.06&93.02\cr 
	\bottomrule
	\end{tabular}
	\caption{Experimental results of ablation study on components of HyCxG.}
	\label{tab:experiment-ablation-component}
\end{table}

%% file: Algorithms/algo-med.tex
\begin{algorithm}[t]
    \fontsize{10.1}{12}\selectfont
    \SetAlgoLined
    \DontPrintSemicolon
	\caption{Compute MD between constructions}
	\label{algo:med}
	\KwIn{The set of constructions $\mathcal{G}$.}
	\KwOut{$\operatorname{MD}$ matrix for $\mathcal{G}$.}
	
	\SetKwFunction{Compute}{Compute\_md}
	\SetKwFunction{Match}{Is\_Match}
    \SetKwProg{Fn}{Function}{:}{}
	
	$\mathcal{G}$ = $\left[g_1, g_2, \cdots, g_n \right]$\;
	Initialize $\operatorname{MD} \in \mathbb{R}^{n \times n}$ matrix with $0$ \;
	\For{$i=0$ to $n-1$}{
	    \For{$j=0$ to $n-1$}{
    		\eIf{$i = j$}{
    			$\operatorname{MD}$[$i$,$j$]$\gets$0\;
    		}
    		{
    		    $\operatorname{MD}$[$i$,$j$]$\gets$\Compute{$g_i$, $g_j$}\;
    		}
		}
	}
	\textbf{return} MD\;
	\;
    \Fn{\Compute{$A$, $B$}}{
        $A=\left[c_1, c_2, \cdots, c_p \right]$, $B=\left[e_1, e_2, \cdots, e_q \right]$\;
        Initialize $\operatorname{dp} \in \mathbb{R}^{(p+1) \times (q+1)}$ matrix with $0$ \;
        \For{$i=1$ to $p+1$}{
            \For{$j=1$ to $q+1$}{
                \eIf{\Match{$A$[$i$-1], $B$[$j$-1]}}{
                    d$\gets$0\;
                }
                {
                    d$\gets$1\;
                }
                $\operatorname{dp}$$\gets$$\min$($\operatorname{dp}$[$i$-1][$j$]+1,$\operatorname{dp}$[$i$][$j$-1]+1, $\operatorname{dp}$[$i$-1][$j$-1]+d)
            }
        }
        dist$\gets \operatorname{dp}$[$m$][$n$] \;
        \textbf{return} dist \;
    }
    \textbf{end function}\;

\end{algorithm}

%% file: Tables/exp-constructicon.tex
\begin{table}[ht]
\fontsize{10}{12}\selectfont
\begin{tabular}{p{1.8cm} p{5cm}}
\toprule
\textbf{Relation} & \textbf{Example} \\ \midrule
\multirow{3}{*}{Polysemy}  & think--PRON--AUX \cr 
                           & believe--PRON--AUX \cr 
                           & know--PRON--AUX \cr \hline
\multirow{3}{*}{Subpart}   & SCONJ--PRON--AUX--VERB--to \cr
                           & SCONJ--PRON--AUX--VERB \cr
                           & SCONJ--PRON--AUX \cr \hline
\multirow{3}{*}{Instance}  & SCONJ--PRON--AUX \cr
                           & if--PRON--AUX \cr
                           & if--PRON--can \cr

\bottomrule 
\end{tabular}

\caption{Examples for three types of inheritance relationships in \emph{constructicon}.}

\label{tab:appendix-constructicon}
\end{table}

%% file: Figures/appendix-constructicon.tex
\begin{figure}[t]
	\centering
	\includegraphics[width=\columnwidth]{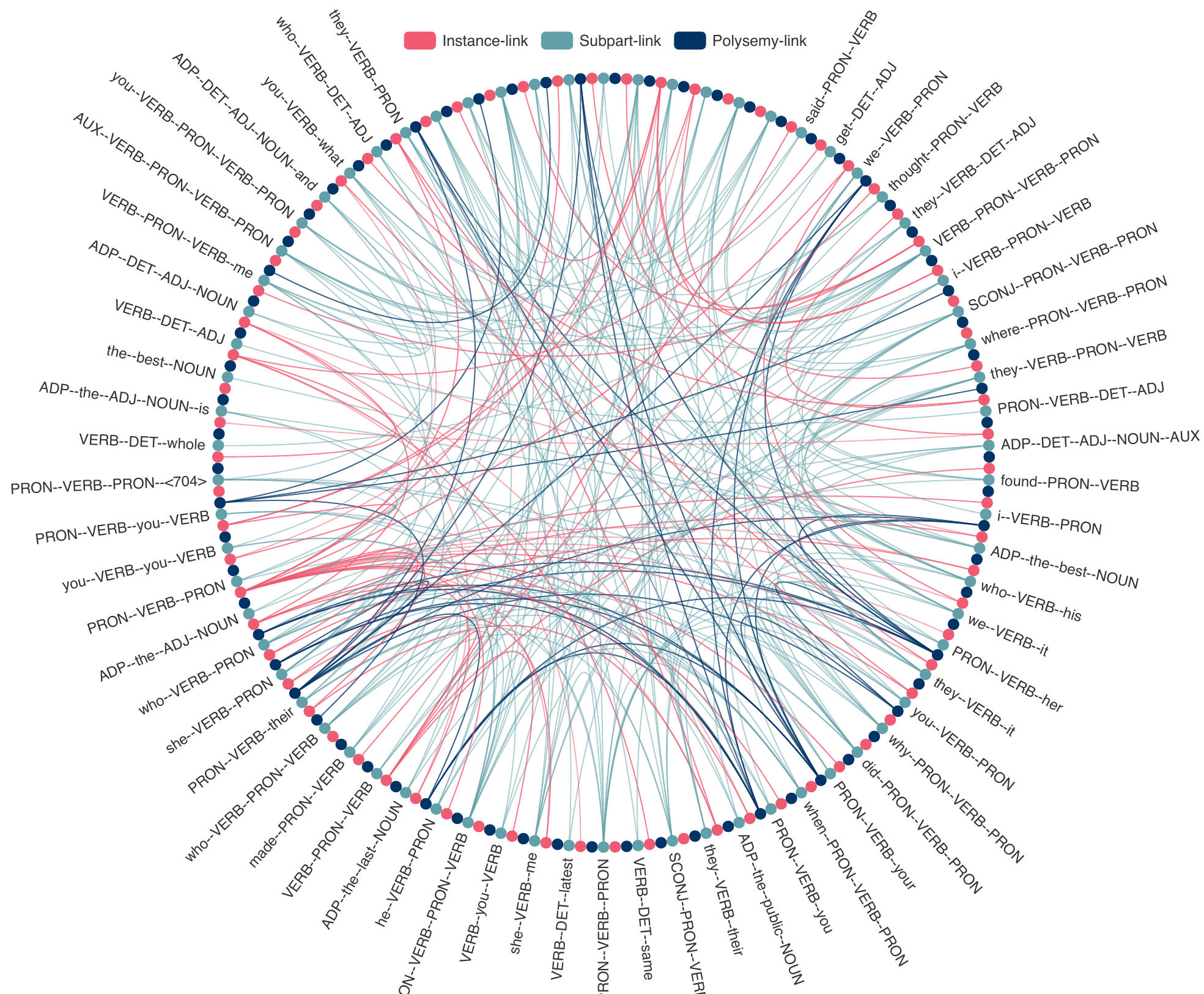} 
	\caption{An example of a sub-network with three inheritance relationships with our visualization tool.}
	\label{fig:appendix-constructicon}
\end{figure}

%% file: Tables/exp-validation-absa.tex
\begin{table}[t]
    \fontsize{10}{11}\selectfont
	\centering
	\begin{tabular}{p{1.7cm} p{0.95cm}<{\centering} p{0.95cm}<{\centering} p{0.95cm}<{\centering}p{0.95cm}<{\centering}}
		\toprule
	    \textbf{Model}& \textbf{Rest14} & \textbf{Lap14} & \textbf{Rest15} & \textbf{Rest16}  \cr
		\midrule
		BERT-SPC                 &84.91&77.59&81.55&90.26\cr 
		R-GAT                    &85.54&78.53&84.13&90.90\cr 
	    KumaGCN                  &85.09&79.00&83.76&91.07\cr
		DGEDT                    &85.27&78.67&82.84&90.42\cr 
		DualGCN                  &85.44&78.84&83.95&91.23\cr \hline
		\textbf{\rule[0pt]{0pt}{11pt}HyCxG} &\textbf{85.98}&\textbf{79.31}&\textbf{84.32}&\textbf{91.88}\cr 
	\bottomrule
	\end{tabular}
	\caption{Experimental results on aspect-based sentiment analysis datasets with validation set.}
	\label{tab:experiment-validation-absa}
\end{table}